\pdfoutput=1
\documentclass{article}
\usepackage{iclr2027_conference,times}

\usepackage{amsmath,amsfonts,bm}

\def\eqref#1{equation~\ref{#1}}

\def\1{\bm{1}}

\DeclareMathAlphabet{\mathsfit}{\encodingdefault}{\sfdefault}{m}{sl}
\SetMathAlphabet{\mathsfit}{bold}{\encodingdefault}{\sfdefault}{bx}{n}

\usepackage{hyperref}
\usepackage{url}
\usepackage{booktabs}
\usepackage{graphicx}
\usepackage{float}
\usepackage{placeins}
\usepackage{enumitem}
\usepackage{xcolor}
\hypersetup{colorlinks=true, linkcolor=red!50!black, citecolor=blue!50!black, urlcolor=magenta!60!black}
\usepackage{amsmath}
\usepackage{microtype}
\usepackage{pifont}
\usepackage{iftex}
\ifPDFTeX
  \usepackage[utf8]{inputenc}
\fi
\usepackage{fvextra}
\ifPDFTeX
  \newcommand{\promptfontfamily}{cmtt}
  \DeclareUnicodeCharacter{2192}{\ensuremath{\rightarrow}}
  \DeclareUnicodeCharacter{2212}{\ensuremath{-}}
\else
  \usepackage{fontspec}
  \newfontfamily\promptmonofont[NFSSFamily=promptmono]{lmmono10-regular.otf}
  \newcommand{\promptfontfamily}{promptmono}
\fi
\newif\ifinprompt

\ifPDFTeX
  \DeclareUnicodeCharacter{2014}{\ifinprompt-{}-\else\textemdash\fi}
\fi
\usepackage{multicol}
\usepackage[breakable]{tcolorbox}
\newtcolorbox{skillhandbook}[1]{
  colback=gray!5,
  colframe=gray!50,
  colbacktitle=gray!15,
  coltitle=black,
  boxrule=0.5pt,
  arc=2pt,
  boxsep=0pt,
  left=8pt,
  right=8pt,
  top=6pt,
  bottom=6pt,
  title={\textbf{#1}},
  title after break={\textbf{#1 (continued)}},
  fonttitle=\small,
  fontupper=\small,
  before upper={\setlength{\parindent}{0pt}},
  breakable
}
\newtcolorbox{methodbox}[1]{
  colback=gray!5,
  colframe=gray!50,
  colbacktitle=gray!15,
  coltitle=black,
  boxrule=0.5pt,
  arc=2pt,
  boxsep=0pt,
  left=8pt,
  right=8pt,
  top=5pt,
  bottom=5pt,
  title={\textbf{#1}},
  fonttitle=\small,
  fontupper=\small,
  before upper={\setlength{\parindent}{0pt}}
}

\usepackage{tabularx}
\usepackage{longtable}
\usepackage{array}
\definecolor{familyGPT}{HTML}{2166AC}
\definecolor{familyClaude}{HTML}{8C3B80}
\definecolor{familyGemini}{HTML}{267346}
\definecolor{familyQwen}{HTML}{7154A1}
\definecolor{familyKimi}{HTML}{C62828}
\definecolor{familyGLM}{HTML}{B45B16}
\definecolor{familyDeepSeek}{HTML}{A68A00}
\definecolor{familyGrok}{HTML}{4B5563}
\newcommand{\modelname}[2]{\textcolor{#1}{\textbf{#2}}}
\newcommand{\scratchpadon}{\textcolor{green!45!black}{\ding{51}}}
\newcommand{\scratchpadoff}{\textcolor{red!75!black}{\ding{55}}}
\newcommand{\dr}[1]{\rlap{\ifcase#1\or*\or**\or***\fi}}

\newcommand{\NumLeaderboardModels}{18}                       %
\newcommand{\GridSize}{$32\times32$}
\newcommand{\NumMaps}{five}
\newcommand{\NumBots}{four}

\newcommand{\ScratchpadCap}{2{,}000}
\newcommand{\TeacherModel}{GPT-5.6 Sol} %
\newcommand{\CodexVersion}{0.146.1}

\newcommand{\CodeURL}{\url{https://github.com/tiptreesystems/textcraft}}

\title{\textsc{TextCraft}: A Text-Native Strategy Benchmark and Training Ground for Language Models}

\author{%
  Alejandro H.~Artiles\textsuperscript{1,2*\dag}\quad
  Amit Sinha\textsuperscript{3,4*\dag}\quad
  Martin Weiss\textsuperscript{4}\\
  \textbf{Anirudh Goyal\quad
  Nasim Rahaman\textsuperscript{4}}\\[3pt]
  \textsuperscript{1}Max Planck Institute for Human Development, Berlin\\
  \textsuperscript{2}Max Planck Institute for Intelligent Systems, T\"ubingen\\
  \textsuperscript{3}McGill University\quad
  \textsuperscript{4}Tiptree Systems\\
  \textsuperscript{*}Equal contribution%
}
\iclrfinalcopy

\begin{document}
\maketitle
\begingroup
\renewcommand{\thefootnote}{\dag}%
\footnotetext{Work done during an internship at Tiptree Systems.}%
\endgroup
\lhead{Preprint}

\begin{abstract}
AI agents are increasingly acting in the real world, managing resources and deciding how to respond as situations unfold. Yet real-world decision-making rarely happens in isolation or with full visibility: it requires balancing long-term planning against short-term execution, reasoning under uncertainty, and adapting to a changing environment and other agents. Competitive games offer a natural, non-saturating way to measure these capabilities, but most language-model benchmarks still rely on static tasks that saturate as models improve, while existing multi-agent games either delegate tactical control to scripts, restrict a model to a single unit, or bury strategic reasoning under complex human interfaces. We introduce \textsc{TextCraft}, an open, text-native real-time strategy environment built around this exact combination of demands: a single model runs an entire side, making both long-term economic and unit-level tactical decisions under partial observability, simultaneous moves, and an opponent that actively reacts. Its rules are simple enough that weak and frontier models fit on one scale, yet rich and customizable enough to remain discriminative as models improve, all in a self-contained Python engine that runs on a CPU for the cost of the model calls alone.
In a 900-game tournament of \NumLeaderboardModels{} models, win rates span 4\% to 96\%, although every model we quiz, from first place to last, reads the text interface with 96--100\% accuracy. Beyond ranking models, \textsc{TextCraft} lets us study interesting behaviors in agents and teams of agents. For example, we show that a strategy can be handed to a weak model, but the ability to adapt it cannot: natural-language handbooks distilled from a weak model's own games, with no weight updates, let Claude Haiku 4.5 and GPT-5.6 Luna beat mid-ranked opponents they never trained against, whereas a handbook written from the rules alone wins nothing, and no handbook overcomes the most adaptive opponent. Communication helps a team only when its members agree on a good plan: in 2v2 play, chat hurts our strongest roster until a single costly agreement is edited out of their conversation, which turns the same channel into an advantage. Because difficulty comes from the opponent and new rules or mechanics are easy to add, \textsc{TextCraft} is designed to grow with the models' capabilities.
\end{abstract}

\begin{center}
All interactive replays and full traces available here: 

\href{https://textcraft-rts.com/}{https://textcraft-rts.com/}

Code: \CodeURL
\end{center}

\section{Introduction}
\label{sec:intro}

\begin{figure}[t]
\centering
\includegraphics[width=\linewidth]{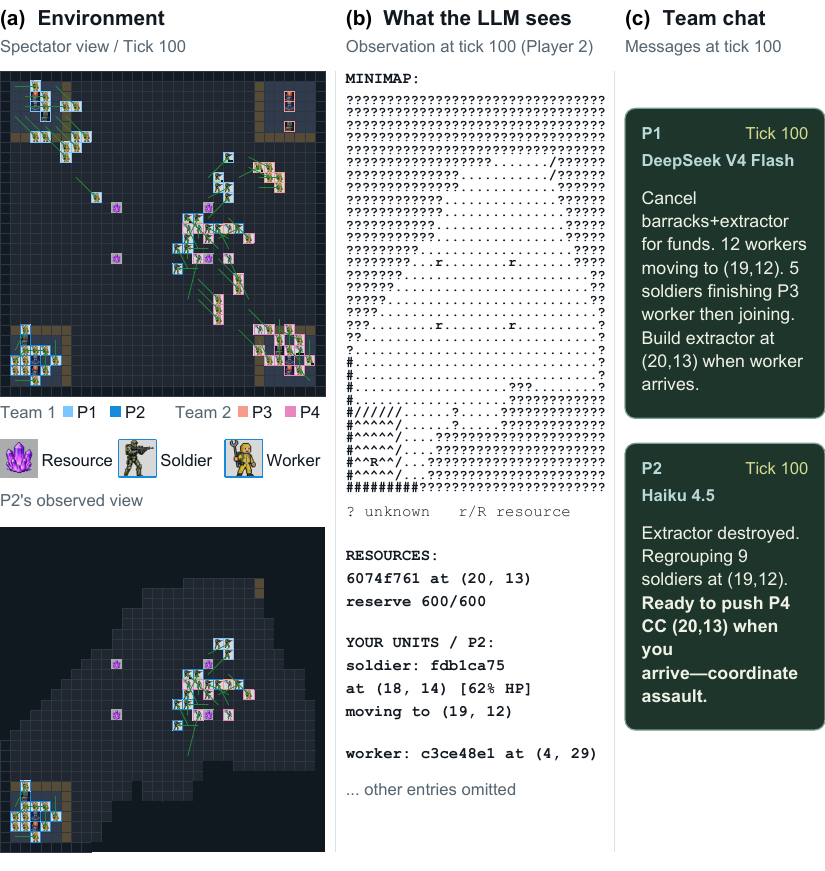}
\caption{
\textbf{\textsc{TextCraft} combines economy, combat, planning under partial observation, and team coordination through text observations, language messages, and structured actions.} (a) The full spectator state and P2's observed view at tick 100, with sprite examples. (b) P2's exact ASCII terrain map and selected resource and unit entries; \texttt{?} marks unknown terrain. Models receive text only; omitted entries are marked, and Appendix~\ref{app:interface} prints the complete observation. (c) Team-chat messages at tick 100, in which Haiku 4.5 and DeepSeek V4 Flash coordinate an assault. Replays and full traces for all 3,615 games are available online.}
\label{fig:environment-overview}
\end{figure}

Most language-model benchmarks test agents against fixed answer keys. They
saturate as models improve, and they reveal little about whether an agent can
anticipate an opponent, adapt a plan over hundreds of decisions, or coordinate
with a teammate under incomplete information. Competitive games offer an
adaptive alternative in which the opponent itself sets the difficulty, but
existing strategy games such as StarCraft~II were designed for human eyes and
hands. Prior LLM interfaces to them either delegate tactical control to scripts
or expose low-level action spaces that obscure strategic
reasoning~\citep{ma2023llmsc2,shao2024swarmbrain,li2024llmpysc2}.

We introduce \textsc{TextCraft}, an open, text-native real-time strategy
environment built specifically for language-model agents. On a \GridSize{} grid
under partial observability, players simultaneously manage an economy, a technology tree,
and an army across hundreds of ticks, receiving structured text observations
and issuing JSON orders (Figure~\ref{fig:environment-overview}). Alongside 1v1
competition, \textsc{TextCraft} supports 2v2 matches in which allies hold
private observations and coordinate through free-text chat. All models run
through a shared, instrumented scaffold that lets us vary reasoning effort,
persistent scratchpads, experience-distilled handbooks, and team communication.

\textbf{Why this design.}
Existing multi-agent benchmarks each isolate a slice of strategic behavior:
shifting alliances, open-ended cooperation, or writing scripts between rounds
(\S\ref{sec:related}). \textsc{TextCraft} combines five properties in one
environment. First, \emph{one model owns the whole stack}: it runs an entire
side at every tick, from the economy to per-unit orders, instead of
controlling a single avatar or offloading micro-control to scripts. Second,
\emph{decisions are made under uncertainty}: fog of war and simultaneous moves
force models to scout, infer an opponent's plan, and commit resources before
they can be sure. Third, \emph{the opponent reacts}: every strategy is tested
against an active adversary, which, as our experiments show, separates
executing a fixed plan from genuine adaptation. Fourth, \emph{weak and strong
models fit on one scale}: the rules and the text interface are simple enough
that all \NumLeaderboardModels{} tested models, from the weakest to the
strongest, play through the same interface without hitting an interface wall.
Finally, \emph{it is cheap, self-contained, and extensible}: the engine is
plain Python on a CPU, so a game costs essentially the model calls
(\$0.23--\$6.93 per model), and new maps, units, or rules can be added without
modifying a heavy game engine.

\textbf{Contributions and main findings.}

\begin{enumerate}[leftmargin=*,itemsep=3pt,topsep=4pt]

\item \textbf{A benchmark whose difficulty comes from the opponent.}
\textsc{TextCraft} combines an economy, tactical combat, partial observability, and team
play in a text-native game. We release the environment, its maps and scripted
opponents, and all game trajectories (\S\ref{sec:env}--\S\ref{sec:protocol}).

\item \textbf{Models read the game equally well and play it very differently.}
In a 900-game tournament of \NumLeaderboardModels{} models, the top model wins
96 of its 100 games and the last wins 4, although both answer almost every
question about an observation correctly. More reasoning helps most models; a
persistent scratchpad rarely does (\S\ref{sec:leaderboard}).

\item \textbf{A strategy can be handed to a model, but the ability to adapt it cannot.}
Skills that a stronger model distills from a weak model's own games lets
it beat mid-ranked models that it never trained against, while a handbook
written from the rules alone wins no game. Against Opus, which adapts to its
opponent, the same skill set win almost nothing (\S\ref{sec:skills}).

\item \textbf{Communication helps when teammates agree on a good plan.}
Chat improves one team's results and worsens another's. The strongest team
loses with chat because its members agree to save for an expansion while
their armies fall behind. When that agreement is removed from their
conversation, they keep producing, attack together, and win all three
replayed games (\S\ref{sec:coop}).

\end{enumerate}

\section{The \textsc{TextCraft} Environment}
\label{sec:env}

\textsc{TextCraft} is a text-native strategy game played on a \GridSize{} grid
in 1v1 or 2v2 matches, with the goal of destroying all opposing command
centers.\footnote{The rules and results in this paper are those of
\textsc{TextCraft} version 1.0.0. The current release, 1.3.2, adds mechanics
(automatic fire for idle and holding units, defensive fire from command
centers and workers, and orbital strikes that a scout must designate and the
opponent can stop) and rebalances units and technology to make play more
diverse and deeper. The changelog in the repository lists every change.} It
probes whether LLMs can reason strategically: weighing trade-offs,
adapting to new information, and responding to an opponent's shifting tactics,
under resource constraints and incomplete information.
\label{sec:interface}
At each discrete \emph{tick}, every player receives a text observation (an
ASCII minimap, visible units and structures, resource counts, production
queues, and feedback on earlier actions; Figure~\ref{fig:environment-overview}b)
and simultaneously submits a JSON array of orders. Orders persist until
completed or replaced, and the engine resolves pathing and damage. The model
decides how closely to direct combat: it can assign each unit a target, or
order units to hold a position, where they fire on enemies that come into
range.

\textbf{The strategic tension.}
Every tick forces a trade-off between economic growth, increasing the army, and immediate survival.
Workers gather finite resources, construct buildings, and repair structures,
but every worker or expansion delays military production. Combat revolves
around three tier-1 infantry types with cyclic advantages, shaped by elevation
bonuses, terrain chokepoints, and damage variance. All of this unfolds under
\emph{fog of war}: unexplored terrain is hidden, and previously spotted enemies
remain only as stale, timestamped records. Scouting resolves that uncertainty,
but risks losing units.

\textbf{Team play and experimental scaffold.}
\label{sec:scaffold}
In 2v2 matches, allies manage separate economies and armies with private
observations; optional free-text chat lets them share sightings or coordinate
attacks without sharing vision
(Figure~\ref{fig:environment-overview}c). Every model plays through the same
custom scaffold, which states the rules and gives no strategic hints, allowing
controlled ablations of reasoning effort, persistent scratchpads, skill
handbooks (\S\ref{sec:skills}), and chat (\S\ref{sec:coop}). The environment
ships \NumMaps{} symmetric 1v1 maps, three of which also have 2v2 versions, and
\NumBots{} scripted opponents (Random, Economy, Rush, and Defensive) that
serve as reference and training opponents. Full unit statistics, game
constants, the maps, and a complete observation are in
Appendix~\ref{app:rules}.

\section{Evaluating Strategic Performance}
\label{sec:protocol}

We evaluate how well language models play \textsc{TextCraft} in two stages.
To compare models fairly, we first select the strongest configuration we find
for each one, so that every model competes at its best. We then compare the
selected agents in a tournament.

\subsection{Evaluation Protocol}

\textbf{Configuration selection.}
\label{sec:seats}
Because performance depends on both the model and its scaffold, we first run
sequential head-to-head brackets to select each model's best thinking level
and whether to enable a persistent scratchpad, in which an agent keeps its own
notes between ticks (protocol and token-cap screening in
Appendix~\ref{app:final-results}). The highest tested reasoning effort wins
for 14 of \NumLeaderboardModels{} models, whereas a persistent scratchpad
helps only three, suggesting that extra per-turn deliberation pays off far
more reliably than carrying self-written notes across ticks.
Table~\ref{tab:selected-configs} lists the selected configurations. We refer
to models by short names throughout: Astra (GPT-6 Astra), Fable
(Claude Fable~5.1), Opus (Claude Opus~5), Sonnet (Claude Sonnet~5), Haiku
(Claude Haiku~4.5), and Sol and Luna (GPT-5.6 Sol and Luna).

\textbf{Tournament.}
The selected configurations compete in ten Swiss rounds. Models are paired by
accumulated tournament points without repeated opponents. Each pairing plays
ten games: one on each of the \NumMaps{} symmetric maps from both starting positions.
Player-relative coordinates and seat swaps reduce positional effects. Every
model faces ten opponents and plays 100 games, producing 900 games overall.
Matches have a 250-tick limit; wins earn one point and draws half a point.

\subsection{Leaderboard and Behavioral Analysis}
\label{sec:leaderboard}

\begin{figure}[!htbp]
\centering
\includegraphics[width=\linewidth]{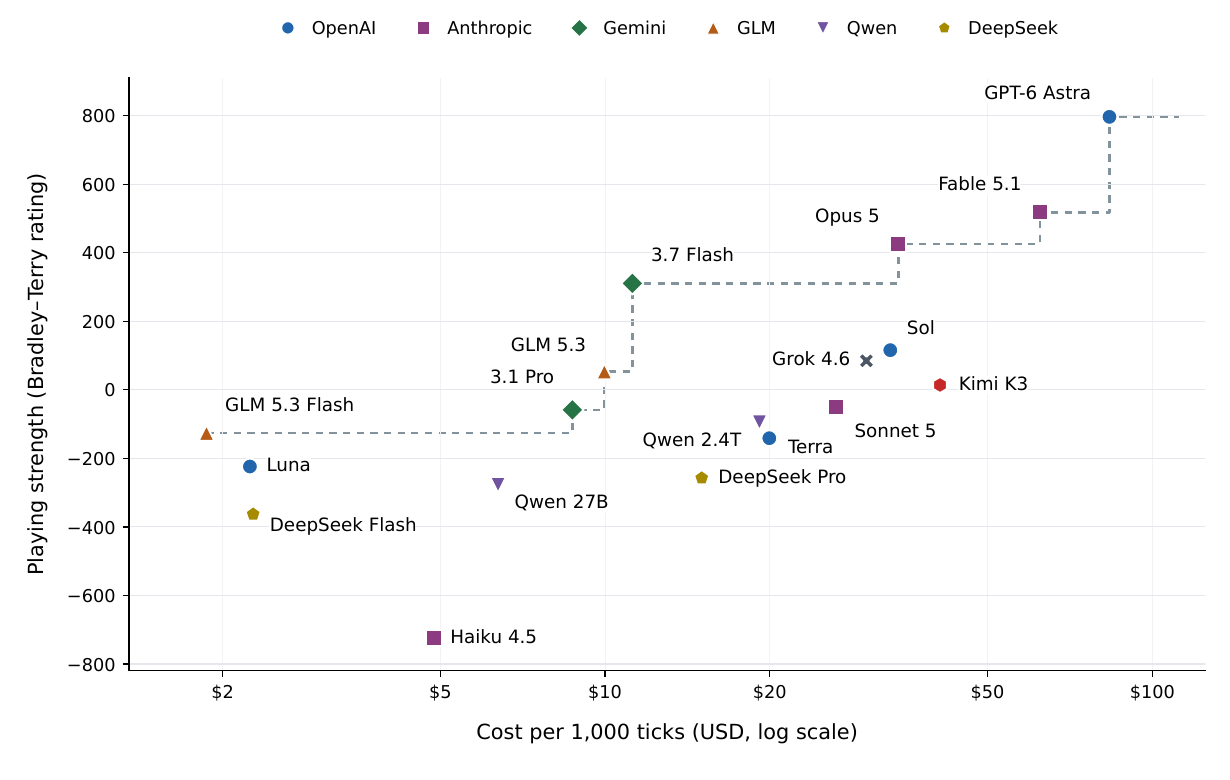}
\caption{\textbf{Competitive performance and cost.} Zero-centered Bradley--Terry rating versus mean cost per 1,000 player ticks across 900 games, with 100 games per model. The cost axis is logarithmic; colors and shapes identify model families. The dashed staircase marks the observed Pareto frontier: the highest rating available at each cost budget.}
\label{fig:rating-cost}
\end{figure}

We rank the selected configurations by tournament record and Bradley--Terry
rating \citep{bradley1952rank}, which estimates relative strength while
accounting for opponents faced, and pair each rating with the monetary cost
per player tick (Figure~\ref{fig:rating-cost}). Costs combine recorded charges with estimates where
necessary, and inference latency is excluded from game outcomes. The resulting
leaderboard therefore compares models under their \emph{selected scaffolds},
not under identical inference budgets. We then analyze economic, military,
exploration, and order-level traces to characterize how models play. Rating
calculations, cost accounting, complete records, and pairwise results appear
in Appendix~\ref{app:final-results}.

\textbf{Competitive performance and cost.}
Astra leads the \NumLeaderboardModels{}-model comparison by Bradley--Terry rating, winning 96 of its 100 games. Fable, Opus, and Gemini~3.7 Flash follow in that order, with 82, 76, and 66 wins, respectively (Figure~\ref{fig:rating-cost}). Astra also has the highest mean cost, at \$0.083 per player tick. Gemini~3.7 Flash, at \$0.011 per tick, roughly a third of Opus's cost and a seventh of Astra's, ranks fourth and offers the best performance-cost trade-off. GLM~5.3 Flash has the lowest measured cost (\$0.0019 per tick) yet ranks twelfth, ahead of six more expensive models, including GPT-5.6~Terra and DeepSeek~V4~Pro.

\textbf{Models read the game equally well and win through play.}
A ranking on a text-only game could reflect how well a model parses dense text
instead of how well it plays. We therefore gave nine models, from first place
to last, a 200-question quiz built from 40 real game observations, with answers
checked against the engine's recorded state (Appendix~\ref{app:obs-quiz}). All
nine answer 96--100\% of the questions correctly, including questions that need
one step of inference over the map. Yet Astra wins 96 of its 100 tournament
games, Gemini~3.1 Pro 45, and Haiku 4. The models read the game almost equally
well; what separates them is what they do with it. Nor do the strongest
models win by exploiting the engine: an audit of the 270 games played by the
three top-ranked models, with scripted checks over all 900 games, found no
rule violation and no win that relies on unintended engine behavior
(Appendix~\ref{app:replay-audit}).

\begin{figure}[t]
\centering
\includegraphics[width=\linewidth]{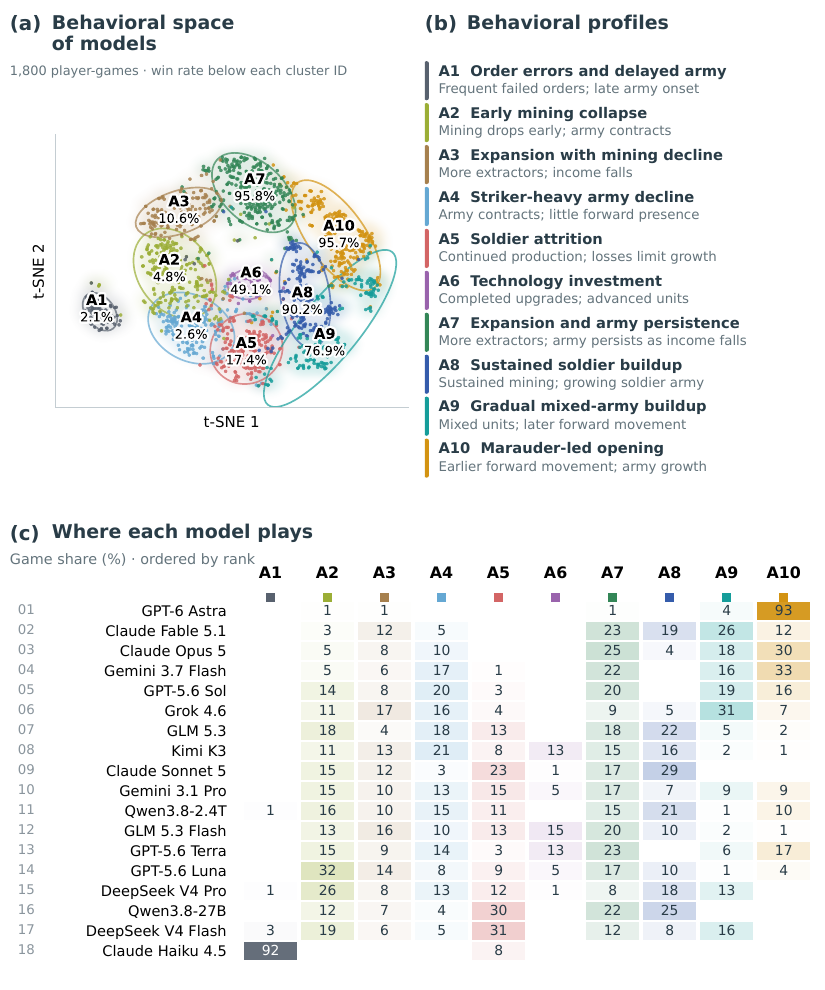}
\caption{\textbf{Behavioral analysis across the leaderboard.}
\textbf{(a)} t-SNE analysis on leaderboard games, colored by clusters fitted to full-game summary and temporal features. Percentages indicate cluster win rates.
\textbf{(b)} Behavioral profiles describe common behavior patterns of clustered groups.
\textbf{(c)} Behavioral profile distribution (\%) of each model's 100 games; blank cells denote zero.}
\label{fig:swiss-strategy-performance}
\label{fig:swiss-tsne}
\end{figure}

\textbf{Model behavioral analysis.}
\label{sec:strategic-patterns}
To look beyond win rates, we cluster all 1,800 tournament player-games
($k$-means, $k=10$) using full-game and temporal features spanning economy,
production, unit composition, combat, and positioning, and project them into
2D with t-SNE \citep{vandermaaten2008tsne}
(Figure~\ref{fig:swiss-strategy-performance}a; feature definitions in
Appendix~\ref{app:swiss-clustering}). Scripted bots mostly lie outside the LLM
reference range (Figure~\ref{fig:swiss-bot-distance}).

The features summarize a whole game, so each of the ten resulting profiles describes how a game unfolded, including how it ended, and not a strategy chosen in advance (Figure~\ref{fig:swiss-strategy-performance}b). Outcome therefore separates the profiles. Four of them (A1--A4) describe a collapse, such as failed orders with a delayed army or an early loss of mining, and win only 2--11\% of their games. Four others (A7--A10) describe different ways of winning and win 77--96\%. What distinguishes the strongest models is which of these winning profiles they reach (Figure~\ref{fig:swiss-strategy-performance}c). Fable spreads most of its games across three of them (A7--A9), whereas Astra plays 93\% of its games in one, a marauder-led opening with sustained forward pressure (A10), and wins all of them. Opponents try to counter with strikers,\footnote{Strikers deal +20\% damage to marauders; marauders deal −20\% to strikers.} but the counter works only when well executed (Appendix~\ref{app:astra-traces}). Astra's four losses all begin with a similar opening: in three, Fable executes the counter well and breaks Astra's pressure (ending in A3 or A9); in the fourth, Gemini~3.7 Flash wins the marauder exchanges with high ground and focus fire (A2). The results thus show several routes to winning and no single optimal strategy, consistent with a balanced game, though other effective strategies may remain unexplored.

\section{Skill Learning Without Weight Updates}
\label{sec:skills}

\textsc{TextCraft} also lets us study how agents improve from experience. We
test whether a language model can learn to play better from its own games,
without weight updates, with the help of a stronger LLM teacher. The teacher
reads the student's games and distills them into natural-language
\emph{skills}, which the student receives in its prompt in later games. The
student learns these skills against four scripted bots on a single map, and we
evaluate whether they transfer to stronger language-model opponents, and to
maps, that never appear in training.

\textbf{Distilling games into a handbook.}
\label{sec:bots}
The student plays the four scripted bots in rotation, one game against each
per generation, and Sol, run as a coding agent in Codex CLI
\citep{openai2025codex}, serves as the teacher. We use Haiku and Luna as
students; they finish 18th and 14th of \NumLeaderboardModels{} in the
tournament. After every generation, the teacher reads all of the student's
games so far, anonymized and unordered, and writes a \emph{skill handbook}
from scratch. This handbook contains at most eight skills, each a named piece of natural-language
advice with a trigger and an action. For example,
Luna's \texttt{skill\_opening\_script} says to build an extractor on the
nearest resource, keep producing and harvesting workers, and add a barracks at
about eight workers (Appendix~\ref{app:skills-handbook}). We rebuild rather
than revise the handbook because we observed empirically that a teacher that revises its previous handbook
anchors on the latest games and regresses after early gains
(Appendix~\ref{app:skills-teacher}). The student receives the handbook with
the rules, and it does use the skills, e.g: ``Following
\texttt{skill\_opening\_economy}, I need to reach 7--8 workers''.

\begin{figure}[!ht]
\centering
\includegraphics[width=\linewidth]{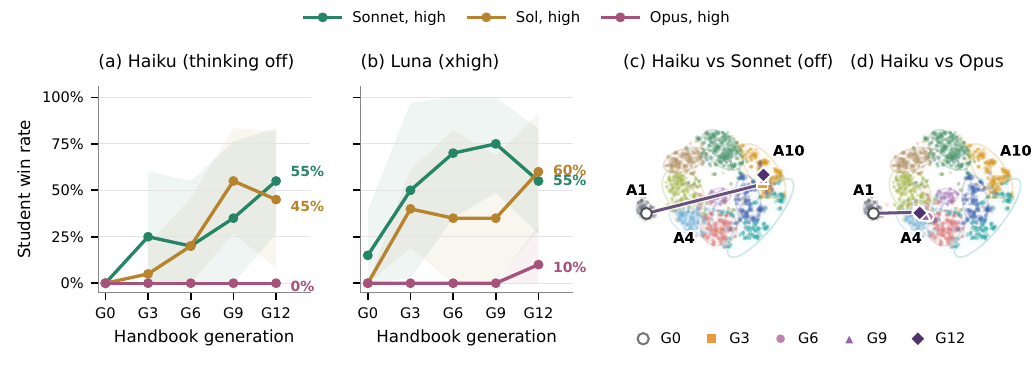}
\caption{
\textbf{Two students take different lessons from the same curriculum.} (a) Haiku with thinking off and (b) Luna with xhigh reasoning against Sonnet, Sol, and Opus with high reasoning; twenty games per checkpoint, shading $\pm1$ SD across ten seed pairs. Both students use the configuration that won their tournament selection, Haiku without thinking and Luna at xhigh reasoning; enabling Haiku's thinking changes its Sonnet results by at most one game (Appendix~\ref{app:skills-haiku-thinking}). (c--d) Haiku's games in the frozen Swiss embedding against different opponents; connected large markers are checkpoint geometric medians. Profiles describe how games unfolded, including how they ended, and not a strategy the student chose.}
\label{fig:skills}
\end{figure}

\textbf{Both students improve against stronger models.}
Each student trains against the bots for twelve generations; we freeze its
handbooks at G3, G6, G9, and G12 and play
twenty games per checkpoint against
Sonnet, Sol, and Opus, none of which receives a handbook
(Figure~\ref{fig:skills}, Appendix~\ref{app:skills-transfer}). Evaluation
games never reach the teacher. Both students improve most against the
mid-range models. Haiku, which lost all ten of its tournament games against
Sonnet, wins 55\% of its G12 games against Sonnet with high thinking (and
95\% against Sonnet without thinking; Appendix~\ref{app:skills-haiku-thinking})
and 45--55\% against Sol at G9--G12. Luna rises
from 15\% to 55--75\% against Sonnet at G9--G12 and from 0\% to 60\% against
Sol at G12. Against Opus, Haiku never wins and Luna wins two games.

\textbf{The gain comes from experience.}
A stronger teacher might help simply by giving good advice. To test this, the
same teacher wrote a handbook under the same limits from the rules alone,
without seeing any game. With this rules-only handbook, Haiku wins 0\% of its
games against high-thinking Sonnet, compared with 50\% across seven
experience-distilled handbooks from four independent runs on the same seeds
($p=0.004$; Appendix~\ref{app:skills-controls}). The rules-only handbook covers every
mechanic with hedged advice that the student has to judge for itself, such as
expanding ``only when rich and secure''. The experience-derived handbooks all
commit to the same concrete plan: six to eight workers, a barracks around
tick~20, and an army built around marauders. Our interpretation is that weaker models do better when following a specific plan than with broad advice, which requires knowing when and how to apply it, that is, adaptability.

\textbf{Skills exploit the habits of mid-range models.}
A few skills can beat models far above the student on the leaderboard because
mid-range models play with fixed habits and adapt little. Sonnet opens with
soldiers in almost every game and keeps producing them, so a student that
counters this habit from the start keeps its advantage. The teacher records
whatever beat the training opponent, so the training bot decides which counter
the student learns. The soldier-only Defensive bot teaches marauders, which
outrange and beat soldiers, a lesson that transfers broadly: trained on that
bot alone, Luna wins 79\% of its games against an eight-LLM panel across five
maps, four of them unseen in training. By contrast, the marauder-only Economy
bot teaches strikers: Luna masters that bot in training but wins only 29\% of
panel games, because strikers rarely counters what LLM opponents actually build
(Appendix~\ref{app:skills-regimes}).

\textbf{Two students take different lessons.}
Haiku's handbook converges on the marauder lesson, and its games unfold
differently as a result. Without a handbook, most of them resemble A1, the
profile of games lost to failed orders and a late army. With one, its games
against Sonnet resemble A10, the profile of a marauder-led opening that keeps
its pressure, which is how almost all of Astra's tournament games unfold
(Figure~\ref{fig:skills}c). Luna's orders fail far less often, so its teacher
spends fewer skills on order legality and formatting, as it must for Haiku,
and more on strategy. Luna's handbook conditions production on the enemy army
it observes, strikers against marauders and marauders against soldiers, and
Luna follows it, answering Opus's marauders with strikers.

\textbf{Skill issue.}
A good opening only goes so far against a model that adapts. Opus with high
reasoning beats every Haiku handbook. These games start with the same marauder
opening, but once Opus takes it apart, Haiku has no answer, and in most of
them Opus wins with a marauder opening of its own. They no longer resemble
A10 and instead end up near A4, a profile of armies in decline
(Figure~\ref{fig:skills}d). Luna's adaptive handbook does
slightly better, winning the only two games any student takes from
high-reasoning Opus, both decided well after the opening. A strategy can be
written down and handed to a weak model. The ability to adapt it once an
opponent takes it apart cannot, a gap we call a ``skill issue''.
\section{When Does Communication Improve Cooperation?}
\label{sec:coop}

In a 2v2 game, two teams of two players compete to destroy each other's
command centers. Allies control separate economies and armies and see only
their own observations, but they can exchange natural-language messages
through a team chat (\S\ref{sec:env}). Intuitively, a shared channel
should help allies report enemy sightings and coordinate attacks, especially
for stronger models. Yet coordination is not free: in a game governed by
the razor-thin trade-off between fast economic growth and fast military
pressure, talking can delay decisions, lock teammates into slow plans, or
let individual errors compound across the team.

\textbf{Matched chat and silent teams.}
To isolate the effect of communication, we pit a chat-enabled team against a
silent team with the exact same two-model roster and settings, holding
individual capability fixed. Both sides have private observations and no
scratchpads. Messages sent on one tick arrive on the next, and the chatting
team sees messages from the previous ten completed ticks. We test three
rosters, Haiku with DeepSeek~V4 Flash (beginner), Sonnet with Gemini~3.1 Pro
(intermediate), and Astra with Fable (expert), each playing 24 games across
three four-player maps, balanced over sides and seats
(Appendix~\ref{app:team-protocol}).

\textbf{A counterintuitive result: chat hurts the strongest team.}
Rather than scaling with capability, the benefit of chat inverts at the top
(Table~\ref{tab:teamchat-wld}). Chat helps the intermediate roster (15 wins,
8 losses, 1 draw) and leaves the beginner roster even (12--12), but it
severely hurts the expert roster, which wins only 6 of 24 games and loses
all eight matches on the King of the Hill map. With 24 games per roster, these
records do not mean chat universally helps or hurts a given capability tier,
but the collapse of the strongest team demands an explanation. To unpack
what went wrong, we read the chat logs alongside replays and paired
coordination measures (Appendix~\ref{app:team-chat-behavior}).

\begin{table}[t]
\centering\small
\caption{Chat-enabled team against a silent team with the same roster (24 games per roster).}
\label{tab:teamchat-wld}
\setlength{\tabcolsep}{5pt}
\begin{tabular}{@{}llrrr@{}}
\toprule Roster & Models & Wins & Losses & Draws \\
\midrule
Beginner & \modelname{familyClaude}{Haiku 4.5} + \modelname{familyDeepSeek}{DeepSeek V4 Flash} & 12 & 12 & 0 \\
Intermediate & \modelname{familyClaude}{Sonnet 5} + \modelname{familyGemini}{Gemini 3.1 Pro} & 15 & 8 & 1 \\
Expert & \modelname{familyGPT}{GPT-6 Astra} + \modelname{familyClaude}{Fable 5.1} & 6 & 18 & 0 \\
\bottomrule
\end{tabular}
\end{table}

\textbf{Compounding errors through misreporting.}
The most obvious way chat can backfire is bad information, and in the lower
tiers, that is exactly what separates helpful talk from wasted tempo. The
intermediate team (Sonnet and Gemini~Pro) keeps its messages grounded in
what matters immediately, talking mostly about forces it already has:
requests become timely reinforcements, teammates split pressure across
opponents, and the chatting team spends more time forward and in joint
combat than its silent counterpart. The beginner team uses the same channel
far less reliably, and its mistakes compound: Haiku reports reinforcements
before they exist, and DeepSeek Flash stalls its own army to wait for them.
Checked against the recorded game state, 21 of Haiku's 25 claims that its
barracks was ready were false. Chat improves the beginner team's army
balance, but not its forward presence or joint combat, leaving its record
even.

\textbf{Agreement slows the experts down.}
Yet misreporting cannot explain the expert collapse. Astra and Fable
exchange precise tactical updates, and none of their 33 checked readiness
claims is false. Digging deeper into their transcripts reveals a subtler
cost of coordination: trying to agree slows them down and locks them into
an expensive, sluggish plan. In 15 of their 24 games, the teammates endorse
saving for additional command centers while their armies fall behind; only
one extra command center is ever completed, yet by tick~50 the chatting
expert team has produced less military than its silent opponent in 22 of 24
games. Meanwhile, the silent team, free from negotiating or waiting on a
shared plan, simply builds and attacks faster. Coordinated assaults can also
leave an ally's base exposed. In the few games the chatting experts win,
they keep military production running, abandon slow plans, and balance
pressure with home defense. The beginner team fails on what is said, but the
expert team fails on what is agreed.

\textbf{Surgically removing the slow agreement.}
If the agreement itself is what slows the expert team down, breaking it
should restore their tempo and flip the result. We tested this directly on
King of the Hill, where Astra and Fable agreed to save for an extra command
center in all eight games and lost all eight. Replaying three of those
games with identical seeds and seats, we intervened only on the chat:
whenever a player proposed the expansion, we edited its teammate's reply to
object (``do not save for or build a hill CC'') while keeping the rest of
the message intact. Neither players' orders nor the silent opponent were
touched, and only 16 messages were edited across the three games, all
between ticks~30 and~50.

The intervention completely reversed the outcome: the chatting team won all
3 replayed games, compared with 0 of 8 without it (two-sided Fisher exact
test, $p=0.006$). The tempo shift is visible well before the game ends: by
tick~50, military production matched the silent opponent instead of lagging
by about 40\%, and unspent resources dropped by half
(Appendix~\ref{app:team-chat-followups}). Freed from the slow expansion
plan, the teammates kept pace with the silent team and turned the channel
into a genuine tactical weapon, calling targets by unit ID and timing joint
volleys (co-targeting 30\% of their kills, versus 4\% for the silent team).
Removing the bad agreement was all it took to turn chat from a liability
into an advantage. Channel timing reflects the same tempo trade-off:
allowing a synchronous message exchange before orders are chosen, at the
cost of a second model call per tick, lifts the same team from 0 of 8 to 5
of 8 on King of the Hill.
\section{Related Work}
\label{sec:related}

\textbf{Games for language agents.}
Text worlds, board and card games, and social-deduction benchmarks test
reasoning and interaction in game settings
\citep{wu2023smartplay,paglieri2024balrog,hu2025lmgamebench,costarelli2024gamebench,guertler2025textarena,xu2023werewolf,light2023avalonbench,yao2025spinbench}.
These benchmarks span single-agent tasks and multiplayer games, including
games with hidden information and simultaneous actions.
StarCraft~II \citep{vinyals2017sc2le,samvelyan2019smac,vinyals2019alphastar}
and MicroRTS \citep{ontanon2013microrts} pose complex real-time challenges.
Prior LLM interfaces differ in the granularity of control they expose:
TextStarCraft~II delegates micro-control to scripts \citep{ma2023llmsc2},
while SwarmBrain combines LLM-directed group commands with rule-based unit
behavior \citep{shao2024swarmbrain}. LLM-PySC2 instead exposes the full
low-level action space, including fine-grained unit operations
\citep{li2024llmpysc2}.
\textsc{TextCraft} gives the language model explicit control over per-unit
target selection and grid-level movement through a compact textual interface,
while the engine handles movement and combat execution. This design makes
tactical decisions directly attributable to the agent while simplifying the
interface through which those decisions are expressed.
The StarCraft~II studies emphasize built-in-AI and task-specific evaluations,
although TextStarCraft~II also reports human matches \citep{ma2023llmsc2};
our evaluation centers on competition between language-model agents.
\textsc{TextCraft} combines three properties in a text-native RTS: moves are simultaneous and
tick-based, so deliberation competes with the orders a model must emit in the
same turn; the game is partially observable with a long horizon; and its 2v2
mode combines cooperation and competition in the same match. Its head-to-head
evaluation follows the arena paradigm
\citep{chiang2024chatbotarena,elo1978rating,bradley1952rank}.

\textbf{What the model decides, and what the engine costs.}
Existing benchmarks each target a different slice of multi-agent behavior,
such as shifting alliances in BattleAgentBench \citep{wang2024battleagentbench}
or open-ended cooperation over long episodes in Alem \citep{tessera2026alem}.
Where they differ from \textsc{TextCraft} is in how much of the game a single
model controls and what it takes to run. In BattleAgentBench and Alem, each
model controls a single tank or avatar and never an entire army. In LLM
Skirmish \citep{llmskirmish}, models write a script between rounds, which
evaluates coding ability and removes in-game decision-making. Prior
StarCraft~II interfaces either offload unit control to scripts or expose a
low-level action space that creates a high barrier for weaker models
\citep{ma2023llmsc2,shao2024swarmbrain,li2024llmpysc2}. Reinforcement-learning
platforms such as MAgent2 \citep{zheng2018magent,terry2020magent} and Neural
MMO~2.0 \citep{suarez2023neuralmmo2} assign one unit per agent, have no native
language interface, and omit construction, army production, and (in Neural
MMO) collision, features that require changing their core rules and not just
adding a text wrapper. \textsc{TextCraft} puts the entire side under one
model's control in a lightweight, CPU-only Python engine, keeping both the
interface floor and the compute cost low enough to compare
\NumLeaderboardModels{} models across hundreds of matches.

\textbf{Reasoning, memory, and learning.}
Chain-of-thought prompting can improve static reasoning tasks
\citep{wei2022cot}, although deliberation also has a cost in agentic settings
\citep{cuadron2025overthinking}.
Agent memories and skill libraries preserve useful experience across decisions
or episodes \citep{shinn2023reflexion,park2023generative,wang2023voyager,zhao2023expel,wang2024awm}.
We separately test thinking and within-game scratchpads.
A related line uses language models as optimizers over prompts and programs
\citep{yang2023opro,fernando2023promptbreeder,yuksekgonul2024textgrad,madaan2023selfrefine},
including coding agents that improve their own scaffolds or discover
algorithms \citep{zhang2025dgm,novikov2025alphaevolve}.
Inspired by reusable behavior handbooks
\citep{didolkar2025metacognitive}, our skill teacher reads complete game
records and rewrites a natural-language strategy handbook, with no weight
updates. We use this established paradigm as a probe of what a model can learn
through its prompt. Because a \textsc{TextCraft} opponent can take a written
plan apart, the game separates two abilities that a static task would merge:
how well a model follows the instructions it is given, and how well it adapts
once they stop working.

\textbf{Cooperation and communication.}
Hanabi studies coordination under hidden information \citep{bard2020hanabi},
while Overcooked studies coordination with human partners
\citep{carroll2019overcooked}. Diplomacy combines communication and strategic
competition \citep{bakhtin2022cicero}; emergent
communication also has a long history in multi-agent learning
\citep{lazaridou2020emergent}.
\textsc{TextCraft} brings free-text team communication into a competitive
strategy game where allies must coordinate despite seeing different parts of
the map, letting us measure when communication pays for its bandwidth.

\section{Limitations and Future Work}
\label{sec:discussion}

Our evaluation fixes several experimental choices to keep comparisons clean
across models.

\textbf{What our setup holds fixed.}
Every model plays through the same scaffold, with identical rules, prompts,
observation format, and action interface (\S\ref{sec:scaffold}). While
this isolates model differences, it means the leaderboard ranks
model--scaffold pairs and not models in the abstract; a different prompt,
memory architecture, or observation layout could reorder them. In particular,
we do not evaluate agent harnesses that write code, invoke external tools, or
manage custom memory during play, which would require sandboxing the engine
and enforcing a comparable tool budget per tick. 

Similarly, to keep large-scale evaluation affordable, we enforce a common per-turn output cap and exclude thinking levels that frequently hit it (Appendix~\ref{app:final-results});
models that continue scaling with larger test-time compute budgets may rank
higher without that ceiling. Finally, we benchmark language models against
one another and scripted bots, leaving out reinforcement-learning and human
baselines.

\textbf{Where \textsc{TextCraft} can go next.}
Most importantly, the environment is fully open and customizable: although our experiments focus on 1v1 and 2v2 matches, nothing prevents testing larger or asymmetric formats such as 3v3 or 4v1, adding new units and mechanics, or probing questions beyond pure strategy, such as multi-agent alignment, where maps with neutral civilians can measure how models trade off military advantage against collateral damage.

\section{Conclusion}
\textsc{TextCraft} was built to measure capabilities that static benchmarks
miss: long-term strategy, on-the-fly adaptability, and multi-agent
communication under uncertainty. Across our experiments, the environment
surfaces behaviors that win rates alone would obscure. First, the bottlenecks
in current models are strategic, not perceptual: models that read the game
state with near-perfect accuracy still differ dramatically in how they
balance an economy or press an advantage. Second, \emph{adaptability} is
harder to transfer than strategy: experience-distilled handbooks let weaker
models beat mid-ranked opponents, but they cannot supply the in-game
adjustments needed when a stronger adversary takes a plan apart. Third,
\emph{communication} cuts both ways: giving teammates a chat channel
does not automatically produce cooperation, but instead amplifies whatever
plan they converge on, whether coordinated focus fire or a slow, mutually
endorsed mistake.

Designed natively for language models and built to be extensible, \textsc{TextCraft} is meant to stay relevant as capabilities advance. Difficulty is generated by the opponent rather than a fixed answer key, so the ceiling rises automatically as frontier models improve. We release the engine, baselines, and full game trajectories so that others can study the behaviors that emerge as models grow. We hope the community finds this benchmark useful and joins us in growing it to adapt to new model capabilities and features.

\section*{Acknowledgments}
This work was funded by Tiptree Advanced Systems Corporation. Model access and compute were supported by Google for Startups Cloud Program credits and Microsoft for Startups Azure credits. Alejandro H. Artiles is supported by the Max Planck Artificial Intelligence Network. Martin Weiss is supported by an IVADO Postdoctoral-Entrepreneur fellowship.

\section*{Reproducibility Statement}
The environment, scaffold, scripted opponents, tournament runner, teacher harness, all experiment drivers, match logs, replays, learned handbooks, and analysis scripts are available at \CodeURL. We also provide a website for interactively exploring replays and all game traces. The main experiments use the ruleset of engine version 1.0.0; later releases change some rules, which the changelog documents. The teacher runs a pinned Codex CLI (\CodexVersion) in an isolated home directory (Codex harness is open source). Appendix~\ref{app:rules} lists all game constants and Table~\ref{tab:final-costs} gives the costs for the four main experiments.

\section*{Ethics Statement}
This work involves no human subjects, human annotators, or personal data. The environment is a fictional strategy game, and all game content is original to this work. The main resource cost is LLM API usage and compute; Table~\ref{tab:final-costs} reports the API costs. The learned skills are natural-language game advice, and we do not foresee dual-use or other harmful applications.

\section*{The Use of Large Language Models}
Large language models are the object of study in this work: they play every game, and a coding agent (OpenAI Codex CLI running \TeacherModel{} in Codex CLI version \CodexVersion) acts as the skill teacher; both are described in \S\ref{sec:scaffold} and \S\ref{sec:skills}. Two frontier models also reviewed replays for the audit in Appendix~\ref{app:replay-audit}, and a language model edits the teammates' messages in the intervention of \S\ref{sec:coop}.

Beyond this, we used generative AI tools in the following ways. (i) \emph{Software implementation}: coding agents (Cursor, Claude Code, and OpenAI Codex CLI) were used to implement parts of the game engine and the experiment drivers, including the ablation experiments. (ii) \emph{Analysis and figures}: the same agents wrote the analysis scripts and the code used to generate the figures in the paper. (iii) \emph{Writing}: LLMs were used  for grammar and style corrections of the manuscript text. The research idea, hypotheses, and experimental design are entirely our own, as is the interpretation of results. We have reviewed all AI-assisted work: LLM-generated code and analysis scripts were checked and tested by the authors, and reported results were verified against the underlying experimental outputs. We take full responsibility for the final content of this work.

\bibliography{references,arxiv_references}

\begin{thebibliography}{44}
\providecommand{\natexlab}[1]{#1}
\providecommand{\url}[1]{\texttt{#1}}
\expandafter\ifx\csname urlstyle\endcsname\relax
  \providecommand{\doi}[1]{doi: #1}\else
  \providecommand{\doi}{doi: \begingroup \urlstyle{rm}\Url}\fi

\bibitem[Bard et~al.(2020)Bard, Foerster, Chandar, Burch, Lanctot, Song,
  Parisotto, Dumoulin, Moitra, Hughes, et~al.]{bard2020hanabi}
Nolan Bard, Jakob~N. Foerster, Sarath Chandar, Neil Burch, Marc Lanctot,
  H.~Francis Song, Emilio Parisotto, Vincent Dumoulin, Subhodeep Moitra, Edward
  Hughes, et~al.
\newblock The {Hanabi} challenge: A new frontier for {AI} research.
\newblock \emph{Artificial Intelligence}, 280:\penalty0 103216, 2020.

\bibitem[Bradley \& Terry(1952)Bradley and Terry]{bradley1952rank}
Ralph~Allan Bradley and Milton~E. Terry.
\newblock Rank analysis of incomplete block designs: {I}. {T}he method of
  paired comparisons.
\newblock \emph{Biometrika}, 39\penalty0 (3/4):\penalty0 324--345, 1952.

\bibitem[Carroll et~al.(2019)Carroll, Shah, Ho, Griffiths, Seshia, Abbeel, and
  Dragan]{carroll2019overcooked}
Micah Carroll, Rohin Shah, Mark~K. Ho, Thomas~L. Griffiths, Sanjit~A. Seshia,
  Pieter Abbeel, and Anca Dragan.
\newblock On the utility of learning about humans for human-{AI} coordination.
\newblock \emph{arXiv preprint arXiv:1910.05789}, 2019.

\bibitem[Chiang et~al.(2024)Chiang, Zheng, Sheng, Angelopoulos, Li, Li, Zhang,
  Zhu, Jordan, Gonzalez, and Stoica]{chiang2024chatbotarena}
Wei-Lin Chiang, Lianmin Zheng, Ying Sheng, Anastasios~Nikolas Angelopoulos,
  Tianle Li, Dacheng Li, Hao Zhang, Banghua Zhu, Michael Jordan, Joseph~E.
  Gonzalez, and Ion Stoica.
\newblock {Chatbot Arena}: An open platform for evaluating {LLMs} by human
  preference.
\newblock \emph{arXiv preprint arXiv:2403.04132}, 2024.

\bibitem[Costarelli et~al.(2024)Costarelli, Allen, Hauksson, Sodunke,
  Hariharan, Cheng, Li, Clymer, and Yadav]{costarelli2024gamebench}
Anthony Costarelli, Mat Allen, Roman Hauksson, Grace Sodunke, Suhas Hariharan,
  Carlson Cheng, Wenjie Li, Joshua Clymer, and Arjun Yadav.
\newblock {GameBench}: Evaluating strategic reasoning abilities of {LLM}
  agents.
\newblock \emph{arXiv preprint arXiv:2406.06613}, 2024.

\bibitem[Cuadron et~al.(2025)Cuadron, Li, Ma, Wang, Wang, Zhuang, Liu,
  Schroeder, Xia, Mao, et~al.]{cuadron2025overthinking}
Alejandro Cuadron, Dacheng Li, Wenjie Ma, Xingyao Wang, Yichuan Wang, Siyuan
  Zhuang, Shu Liu, Luis~Gaspar Schroeder, Tian Xia, Huanzhi Mao, et~al.
\newblock The danger of overthinking: Examining the reasoning-action dilemma in
  agentic tasks.
\newblock \emph{arXiv preprint arXiv:2502.08235}, 2025.

\bibitem[Didolkar et~al.(2025)Didolkar, Ballas, Arora, and
  Goyal]{didolkar2025metacognitive}
Aniket Didolkar, Nicolas Ballas, Sanjeev Arora, and Anirudh Goyal.
\newblock Metacognitive reuse: Turning recurring {LLM} reasoning into concise
  behaviors.
\newblock \emph{arXiv preprint arXiv:2509.13237}, 2025.
\newblock URL \url{https://arxiv.org/abs/2509.13237}.

\bibitem[Elo(1978)]{elo1978rating}
Arpad~E. Elo.
\newblock \emph{The Rating of Chessplayers, Past and Present}.
\newblock Arco Publishing, 1978.

\bibitem[Fernando et~al.(2023)Fernando, Banarse, Michalewski, Osindero, and
  Rockt{\"a}schel]{fernando2023promptbreeder}
Chrisantha Fernando, Dylan Banarse, Henryk Michalewski, Simon Osindero, and Tim
  Rockt{\"a}schel.
\newblock Promptbreeder: Self-referential self-improvement via prompt
  evolution.
\newblock \emph{arXiv preprint arXiv:2309.16797}, 2023.

\bibitem[Guertler et~al.(2025)Guertler, Cheng, Yu, Liu, Choshen, and
  Tan]{guertler2025textarena}
Leon Guertler, Bobby Cheng, Simon Yu, Bo~Liu, Leshem Choshen, and Cheston Tan.
\newblock {TextArena}.
\newblock \emph{arXiv preprint arXiv:2504.11442}, 2025.

\bibitem[Hu et~al.(2025)Hu, Huo, Zhang, Yu, Xing, Stoica, Rosing, Jin, and
  Zhang]{hu2025lmgamebench}
Lanxiang Hu, Mingjia Huo, Yuxuan Zhang, Haoyang Yu, Eric~P. Xing, Ion Stoica,
  Tajana Rosing, Haojian Jin, and Hao Zhang.
\newblock {lmgame-Bench}: How good are {LLMs} at playing games?
\newblock \emph{arXiv preprint arXiv:2505.15146}, 2025.

\bibitem[Lazaridou \& Baroni(2020)Lazaridou and Baroni]{lazaridou2020emergent}
Angeliki Lazaridou and Marco Baroni.
\newblock Emergent multi-agent communication in the deep learning era.
\newblock \emph{arXiv preprint arXiv:2006.02419}, 2020.

\bibitem[Li et~al.(2024)Li, Ni, Qi, Jiang, Lu, Xu, Liu, Li, Guo, Ma,
  et~al.]{li2024llmpysc2}
Zongyuan Li, Yanan Ni, Runnan Qi, Lumin Jiang, Chang Lu, Xiaojie Xu, Xiangbei
  Liu, Pengfei Li, Yunzheng Guo, Zhe Ma, et~al.
\newblock {LLM-PySC2}: {StarCraft II} learning environment for large language
  models.
\newblock \emph{arXiv preprint arXiv:2411.05348}, 2024.

\bibitem[Light et~al.(2023)Light, Cai, Shen, and Hu]{light2023avalonbench}
Jonathan Light, Min Cai, Sheng Shen, and Ziniu Hu.
\newblock {AvalonBench}: Evaluating {LLMs} playing the game of {Avalon}.
\newblock \emph{arXiv preprint arXiv:2310.05036}, 2023.

\bibitem[Ma et~al.(2023)Ma, Mi, Zeng, Yan, Wu, Lin, Zhang, and
  Wang]{ma2023llmsc2}
Weiyu Ma, Qirui Mi, Yongcheng Zeng, Xue Yan, Yuqiao Wu, Runji Lin, Haifeng
  Zhang, and Jun Wang.
\newblock Large language models play {StarCraft II}: Benchmarks and a chain of
  summarization approach.
\newblock \emph{arXiv preprint arXiv:2312.11865}, 2023.

\bibitem[Madaan et~al.(2023)Madaan, Tandon, Gupta, Hallinan, Gao, Wiegreffe,
  Alon, Dziri, Prabhumoye, Yang, et~al.]{madaan2023selfrefine}
Aman Madaan, Niket Tandon, Prakhar Gupta, Skyler Hallinan, Luyu Gao, Sarah
  Wiegreffe, Uri Alon, Nouha Dziri, Shrimai Prabhumoye, Yiming Yang, et~al.
\newblock {Self-Refine}: Iterative refinement with self-feedback.
\newblock \emph{arXiv preprint arXiv:2303.17651}, 2023.

\bibitem[McPheeters(2026)]{llmskirmish}
Kai McPheeters.
\newblock {LLM Skirmish}: {Screeps}.
\newblock \url{https://llmskirmish.com/screeps}, 2026.
\newblock Accessed 2026-11-30.

\bibitem[{Meta Fundamental AI Research Diplomacy Team (FAIR)} et~al.(2022){Meta
  Fundamental AI Research Diplomacy Team (FAIR)}, Bakhtin, Brown, Dinan,
  Farina, Flaherty, Fried, Goff, Gray, Hu, et~al.]{bakhtin2022cicero}
{Meta Fundamental AI Research Diplomacy Team (FAIR)}, Anton Bakhtin, Noam
  Brown, Emily Dinan, Gabriele Farina, Colin Flaherty, Daniel Fried, Andrew
  Goff, Jonathan Gray, Hengyuan Hu, et~al.
\newblock Human-level play in the game of {Diplomacy} by combining language
  models with strategic reasoning.
\newblock \emph{Science}, 378\penalty0 (6624):\penalty0 1067--1074, 2022.

\bibitem[Novikov et~al.(2025)Novikov, V{\~u}, Eisenberger, Dupont, Huang,
  Wagner, Shirobokov, Kozlovskii, Ruiz, Mehrabian,
  et~al.]{novikov2025alphaevolve}
Alexander Novikov, Ng{\^a}n V{\~u}, Marvin Eisenberger, Emilien Dupont, Po-Sen
  Huang, Adam~Zsolt Wagner, Sergey Shirobokov, Borislav Kozlovskii, Francisco
  J.~R. Ruiz, Abbas Mehrabian, et~al.
\newblock {AlphaEvolve}: A coding agent for scientific and algorithmic
  discovery.
\newblock \emph{arXiv preprint arXiv:2506.13131}, 2025.

\bibitem[Onta{\~n}{\'o}n(2013)]{ontanon2013microrts}
Santiago Onta{\~n}{\'o}n.
\newblock The combinatorial multi-armed bandit problem and its application to
  real-time strategy games.
\newblock In \emph{Proceedings of the AAAI Conference on Artificial
  Intelligence and Interactive Digital Entertainment (AIIDE)}, pp.\  58--64,
  2013.

\bibitem[{OpenAI}(2025)]{openai2025codex}
{OpenAI}.
\newblock {OpenAI Codex CLI}.
\newblock \url{https://github.com/openai/codex}, 2025.
\newblock Accessed 2026-09-26.

\bibitem[Paglieri et~al.(2024)Paglieri, Cupia{\l}, Coward, Piterbarg,
  Wo{\l}czyk, Khan, Pignatelli, Kuci{\'n}ski, Pinto, Fergus,
  et~al.]{paglieri2024balrog}
Davide Paglieri, Bart{\l}omiej Cupia{\l}, Samuel Coward, Ulyana Piterbarg,
  Maciej Wo{\l}czyk, Akbir Khan, Eduardo Pignatelli, {\L}ukasz Kuci{\'n}ski,
  Lerrel Pinto, Rob Fergus, et~al.
\newblock {BALROG}: Benchmarking agentic {LLM} and {VLM} reasoning on games.
\newblock \emph{arXiv preprint arXiv:2411.13543}, 2024.

\bibitem[Park et~al.(2023)Park, O'Brien, Cai, Morris, Liang, and
  Bernstein]{park2023generative}
Joon~Sung Park, Joseph~C. O'Brien, Carrie~J. Cai, Meredith~Ringel Morris, Percy
  Liang, and Michael~S. Bernstein.
\newblock Generative agents: Interactive simulacra of human behavior.
\newblock \emph{arXiv preprint arXiv:2304.03442}, 2023.

\bibitem[Samvelyan et~al.(2019)Samvelyan, Rashid, Schroeder~de Witt, Farquhar,
  Nardelli, Rudner, Hung, Torr, Foerster, and Whiteson]{samvelyan2019smac}
Mikayel Samvelyan, Tabish Rashid, Christian Schroeder~de Witt, Gregory
  Farquhar, Nantas Nardelli, Tim G.~J. Rudner, Chia-Man Hung, Philip H.~S.
  Torr, Jakob Foerster, and Shimon Whiteson.
\newblock The {StarCraft} multi-agent challenge.
\newblock In \emph{Proceedings of the International Conference on Autonomous
  Agents and MultiAgent Systems (AAMAS)}, 2019.

\bibitem[Shao et~al.(2024)Shao, Jiang, Zuo, and Liu]{shao2024swarmbrain}
Xiao Shao, Weifu Jiang, Fei Zuo, and Mengqing Liu.
\newblock {SwarmBrain}: Embodied agent for real-time strategy game {StarCraft
  II} via large language models.
\newblock \emph{arXiv preprint arXiv:2401.17749}, 2024.

\bibitem[Shinn et~al.(2023)Shinn, Cassano, Berman, Gopinath, Narasimhan, and
  Yao]{shinn2023reflexion}
Noah Shinn, Federico Cassano, Edward Berman, Ashwin Gopinath, Karthik
  Narasimhan, and Shunyu Yao.
\newblock Reflexion: Language agents with verbal reinforcement learning.
\newblock \emph{arXiv preprint arXiv:2303.11366}, 2023.

\bibitem[Su{\'a}rez et~al.(2023)Su{\'a}rez, Isola, Choe, Bloomin, Li,
  Pinnaparaju, Kanna, Scott, Sullivan, Shuman, de~Alc{\^a}ntara, Bradley,
  Castricato, You, Jiang, Li, Chen, and Zhu]{suarez2023neuralmmo2}
Joseph Su{\'a}rez, Phillip Isola, Kyoung~Whan Choe, David Bloomin, Hao~Xiang
  Li, Nikhil Pinnaparaju, Nishaanth Kanna, Daniel Scott, Ryan Sullivan, Rose~S.
  Shuman, Lucas de~Alc{\^a}ntara, Herbie Bradley, Louis Castricato, Kirsty You,
  Yuhao Jiang, Qimai Li, Jiaxin Chen, and Xiaolong Zhu.
\newblock Neural {MMO} 2.0: A massively multi-task addition to massively
  multi-agent learning.
\newblock \emph{arXiv preprint arXiv:2311.03736}, 2023.

\bibitem[Terry et~al.(2020)Terry, Black, and Jayakumar]{terry2020magent}
Jordan~K. Terry, Benjamin Black, and Mario Jayakumar.
\newblock {MAgent}.
\newblock \url{https://github.com/Farama-Foundation/MAgent2}, 2020.
\newblock GitHub repository, maintained as MAgent2 by the Farama Foundation.

\bibitem[Tessera et~al.(2026)Tessera, Szecsenyi, Barker, Rutherford, Paglieri,
  Scannell, Gouk, Crowley, Rockt{\"a}schel, and Storkey]{tessera2026alem}
Kale-ab~Abebe Tessera, Andras Szecsenyi, Cameron Barker, Alexander Rutherford,
  Davide Paglieri, Aidan Scannell, Henry Gouk, Elliot~J. Crowley, Tim
  Rockt{\"a}schel, and Amos Storkey.
\newblock Benchmarking open-ended multi-agent coordination in language agents.
\newblock \emph{arXiv preprint arXiv:2606.08340}, 2026.
\newblock To appear at NeurIPS 2026, Evaluations and Datasets Track.

\bibitem[van~der Maaten \& Hinton(2008)van~der Maaten and
  Hinton]{vandermaaten2008tsne}
Laurens van~der Maaten and Geoffrey Hinton.
\newblock Visualizing data using {t-SNE}.
\newblock \emph{Journal of Machine Learning Research}, 9\penalty0
  (86):\penalty0 2579--2605, 2008.
\newblock URL \url{https://www.jmlr.org/papers/v9/vandermaaten08a.html}.

\bibitem[Vinyals et~al.(2017)Vinyals, Ewalds, Bartunov, Georgiev, Vezhnevets,
  Yeo, Makhzani, K{\"u}ttler, Agapiou, Schrittwieser, et~al.]{vinyals2017sc2le}
Oriol Vinyals, Timo Ewalds, Sergey Bartunov, Petko Georgiev, Alexander~Sasha
  Vezhnevets, Michelle Yeo, Alireza Makhzani, Heinrich K{\"u}ttler, John
  Agapiou, Julian Schrittwieser, et~al.
\newblock {StarCraft II}: A new challenge for reinforcement learning.
\newblock \emph{arXiv preprint arXiv:1708.04782}, 2017.

\bibitem[Vinyals et~al.(2019)Vinyals, Babuschkin, Czarnecki, Mathieu, Dudzik,
  Chung, Choi, Powell, Ewalds, Georgiev, et~al.]{vinyals2019alphastar}
Oriol Vinyals, Igor Babuschkin, Wojciech~M. Czarnecki, Micha{\"e}l Mathieu,
  Andrew Dudzik, Junyoung Chung, David~H. Choi, Richard Powell, Timo Ewalds,
  Petko Georgiev, et~al.
\newblock Grandmaster level in {StarCraft II} using multi-agent reinforcement
  learning.
\newblock \emph{Nature}, 575\penalty0 (7782):\penalty0 350--354, 2019.

\bibitem[Wang et~al.(2023)Wang, Xie, Jiang, Mandlekar, Xiao, Zhu, Fan, and
  Anandkumar]{wang2023voyager}
Guanzhi Wang, Yuqi Xie, Yunfan Jiang, Ajay Mandlekar, Chaowei Xiao, Yuke Zhu,
  Linxi Fan, and Anima Anandkumar.
\newblock {Voyager}: An open-ended embodied agent with large language models.
\newblock \emph{arXiv preprint arXiv:2305.16291}, 2023.

\bibitem[Wang et~al.(2024{\natexlab{a}})Wang, Zhang, Feng, Wang, and
  Tang]{wang2024battleagentbench}
Wei Wang, Dan Zhang, Tao Feng, Boyan Wang, and Jie Tang.
\newblock {BattleAgentBench}: A benchmark for evaluating cooperation and
  competition capabilities of language models in multi-agent systems.
\newblock \emph{arXiv preprint arXiv:2408.15971}, 2024{\natexlab{a}}.

\bibitem[Wang et~al.(2024{\natexlab{b}})Wang, Mao, Fried, and
  Neubig]{wang2024awm}
Zora~Zhiruo Wang, Jiayuan Mao, Daniel Fried, and Graham Neubig.
\newblock Agent workflow memory.
\newblock \emph{arXiv preprint arXiv:2409.07429}, 2024{\natexlab{b}}.

\bibitem[Wei et~al.(2022)Wei, Wang, Schuurmans, Bosma, Ichter, Xia, Chi, Le,
  and Zhou]{wei2022cot}
Jason Wei, Xuezhi Wang, Dale Schuurmans, Maarten Bosma, Brian Ichter, Fei Xia,
  Ed~Chi, Quoc~V. Le, and Denny Zhou.
\newblock Chain-of-thought prompting elicits reasoning in large language
  models.
\newblock \emph{arXiv preprint arXiv:2201.11903}, 2022.

\bibitem[Wu et~al.(2023)Wu, Tang, Mitchell, and Li]{wu2023smartplay}
Yue Wu, Xuan Tang, Tom~M. Mitchell, and Yuanzhi Li.
\newblock {SmartPlay}: A benchmark for {LLMs} as intelligent agents.
\newblock \emph{arXiv preprint arXiv:2310.01557}, 2023.

\bibitem[Xu et~al.(2023)Xu, Wang, Li, Luo, Wang, Liu, and Liu]{xu2023werewolf}
Yuzhuang Xu, Shuo Wang, Peng Li, Fuwen Luo, Xiaolong Wang, Weidong Liu, and
  Yang Liu.
\newblock Exploring large language models for communication games: An empirical
  study on {Werewolf}.
\newblock \emph{arXiv preprint arXiv:2309.04658}, 2023.

\bibitem[Yang et~al.(2023)Yang, Wang, Lu, Liu, Le, Zhou, and
  Chen]{yang2023opro}
Chengrun Yang, Xuezhi Wang, Yifeng Lu, Hanxiao Liu, Quoc~V. Le, Denny Zhou, and
  Xinyun Chen.
\newblock Large language models as optimizers.
\newblock \emph{arXiv preprint arXiv:2309.03409}, 2023.

\bibitem[Yao et~al.(2025)Yao, Wang, Hsieh, Zhou, Zou, Cheng, Wang, and
  Viswanath]{yao2025spinbench}
Jianzhu Yao, Kevin Wang, Ryan Hsieh, Haisu Zhou, Tianqing Zou, Zerui Cheng,
  Zhangyang Wang, and Pramod Viswanath.
\newblock {SPIN-Bench}: How well do {LLMs} plan strategically and reason
  socially?
\newblock \emph{arXiv preprint arXiv:2503.12349}, 2025.

\bibitem[Yuksekgonul et~al.(2024)Yuksekgonul, Bianchi, Boen, Liu, Huang,
  Guestrin, and Zou]{yuksekgonul2024textgrad}
Mert Yuksekgonul, Federico Bianchi, Joseph Boen, Sheng Liu, Zhi Huang, Carlos
  Guestrin, and James Zou.
\newblock {TextGrad}: Automatic ``differentiation'' via text.
\newblock \emph{arXiv preprint arXiv:2406.07496}, 2024.

\bibitem[Zhang et~al.(2025)Zhang, Hu, Lu, Lange, and Clune]{zhang2025dgm}
Jenny Zhang, Shengran Hu, Cong Lu, Robert Lange, and Jeff Clune.
\newblock {Darwin G{\"o}del Machine}: Open-ended evolution of self-improving
  agents.
\newblock \emph{arXiv preprint arXiv:2505.22954}, 2025.

\bibitem[Zhao et~al.(2023)Zhao, Huang, Xu, Lin, Liu, and Huang]{zhao2023expel}
Andrew Zhao, Daniel Huang, Quentin Xu, Matthieu Lin, Yong-Jin Liu, and Gao
  Huang.
\newblock {ExpeL}: {LLM} agents are experiential learners.
\newblock \emph{arXiv preprint arXiv:2308.10144}, 2023.

\bibitem[Zheng et~al.(2018)Zheng, Yang, Cai, Zhou, Zhang, Wang, and
  Yu]{zheng2018magent}
Lianmin Zheng, Jiacheng Yang, Han Cai, Ming Zhou, Weinan Zhang, Jun Wang, and
  Yong Yu.
\newblock {MAgent}: A many-agent reinforcement learning platform for artificial
  collective intelligence.
\newblock In \emph{Thirty-Second AAAI Conference on Artificial Intelligence},
  2018.

\end{thebibliography}
\bibliographystyle{iclr2027_conference}

\clearpage
\appendix
\raggedbottom
\makeatletter\setlength{\@fptop}{0pt}\setlength{\@fpsep}{16pt}\makeatother
\noindent\textbf{Appendix contents.}
\begin{itemize}[leftmargin=2.2em,itemsep=1pt,topsep=3pt]
\item[\textbf{\ref{app:rules}}] Game rules, the eight maps, and a complete observation.
\item[\textbf{\ref{app:final-results}}] Configuration selection, the tournament, and the cost of all experiments.
\item[\textbf{\ref{app:validity}}] Observation-reading quiz and replay audit.
\item[\textbf{\ref{app:swiss-tsne}}] Behavioral profiles of the tournament games, and how Astra's opening is beaten.
\item[\textbf{\ref{app:skills}}] Skill learning: teachers, transfer, training opponents, controls, and the handbooks.
\item[\textbf{\ref{app:team-tsne}}] 2v2 team games: team profiles, what the conversations show, and the message intervention.
\end{itemize}
\noindent Records are wins--losses--draws, and win rates count a draw as a non-win.

\section{Game Rules, Maps, and the Text Interface}
\label{app:rules}
A player or team wins by destroying every opposing command center while keeping at least one of its own; in 2v2, both opponents must be eliminated. A side may also surrender. If both sides are eliminated on the same tick, or the tick limit is reached, the game is a draw. Units move in eight directions. Attack range and vision use Chebyshev distance, $d=\max(|\Delta x|,|\Delta y|)$, so a diagonal neighbor is at distance 1.

\begin{table}[H]
\centering\small
\caption{Units. Cooldown (CD) is the number of ticks between attacks. The tank deals $3\times3$ splash damage and cannot fire at range $<3$.}
\begin{tabular}{cccccccccc}
\toprule
Unit & HP & Dmg & Range & Speed & Vision & CD & Cost & Build & Tier \\
\midrule
worker   & 25 & --  & -- & 2 & 2 & -- & 25  & 2 & 0 \\
scout    & 20 & 3  & 1 & 4 & 6 & 2 & 25  & 2 & 1 \\
soldier  & 40 & 10 & 2 & 2 & 3 & 2 & 50  & 3 & 1 \\
marauder & 30 & 15 & 4 & 2 & 4 & 2 & 60  & 3 & 1 \\
striker  & 35 & 8  & 1 & 3 & 3 & 2 & 40  & 2 & 1 \\
enforcer & 80 & 15 & 1 & 1 & 3 & 3 & 100 & 4 & 2 \\
tank     & 45 & 12 & 6 & 1 & 5 & 4 & 150 & 5 & 2 \\
\bottomrule
\end{tabular}
\end{table}

\begin{table}[H]
\centering\small
\begin{minipage}[t]{0.40\linewidth}
\centering
\caption{Damage modifiers among tier-1 infantry: row attacks column (units only).}
\setlength{\tabcolsep}{4pt}
\begin{tabular}{@{}cccc@{}}
\toprule
 & soldier & marauder & striker \\
\midrule
soldier  & --  & $-40\%$ & $+30\%$ \\
marauder & $+40\%$ & --  & $-20\%$ \\
striker  & $-30\%$ & $+20\%$ & --  \\
\bottomrule
\end{tabular}
\end{minipage}\hfill
\begin{minipage}[t]{0.58\linewidth}
\centering
\caption{Other game constants.}
\label{tab:other-constants}
\begin{tabular}{lc}
\toprule
Parameter & Value \\
\midrule
Damage variance & $\pm25\%$ \\
High-ground bonus & $+25\%$ \\
Starting resources & 200 \\
Starting workers & 2 per command center \\
Worker carry capacity & 8 \\
Cancellation refund & 75\% \\
Worker repair & 5 HP/tick at 1 resource/tick \\
Tick limit (tournament, 2v2) & 250 \\
\bottomrule
\end{tabular}
\end{minipage}
\end{table}

\begin{table}[H]
\centering\small
\caption{Structures and upgrades. Upgrade entries give cost, research time in ticks, and effect.}
\label{tab:research-rules}
\setlength{\tabcolsep}{4pt}
\begin{tabularx}{\linewidth}{@{}ccccc>{\centering\arraybackslash}p{2cm}>{\centering\arraybackslash}X@{}}
\toprule
Structure & HP & Cost & Build & Vision & Prerequisite & Upgrades or ability \\
\midrule
command center & 500 & 400 & 20 & 4 & -- & -- \\
extractor      & 150 & 100 & 5  & 2 & -- & -- \\
barracks       & 200 & 150 & 8  & 3 & -- & -- \\
tech lab       & 180 & 200 & 10 & 2 & barracks & optics (100, 10, +2 vision);\par plating (150, 12, +15\% HP for new units) \\
machine shop   & 180 & 200 & 10 & 2 & barracks & boosters (100, 10, +1 speed);\par ordnance (150, 12, +15\% damage) \\
weapons lab    & 200 & 500 & 15 & 2 & tech lab + machine shop & orbital strike (300; 200 damage in a $3\times3$ area after a 5-tick countdown; 30-tick cooldown) \\
\bottomrule
\end{tabularx}
\end{table}

\subsection{Maps}
\label{app:maps}
\begin{figure}[H]
\centering
\includegraphics[width=\linewidth]{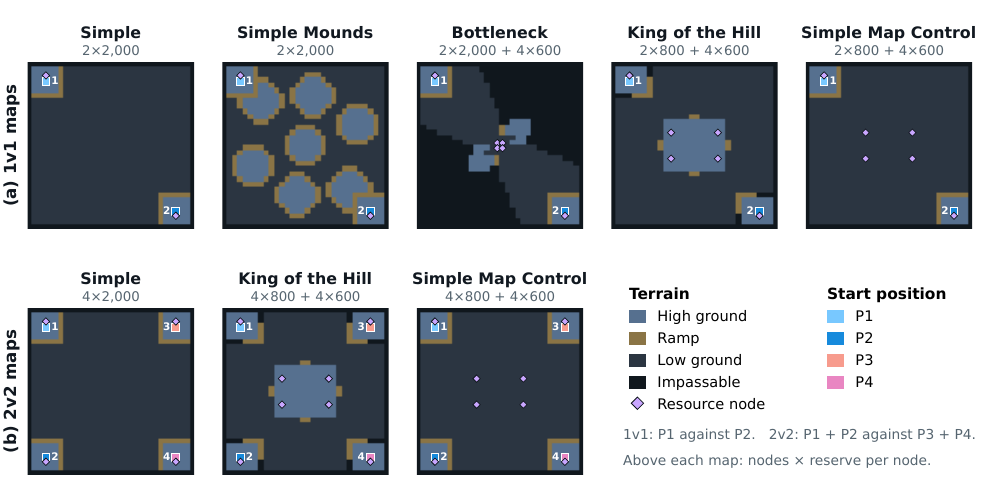}
\caption{\textbf{The evaluation maps.} Terrain, resource nodes, and starting positions of the five 1v1 maps and the three 2v2 maps.}
\label{fig:maps}
\end{figure}

All maps are $32\times32$ (Figure~\ref{fig:maps}). Every player starts on a high-ground plateau next to a resource node. Bottleneck, King of the Hill, and Simple Map Control add contested nodes in the center, and three maps have 2v2 versions in which teammates start on the same side.

\subsection{What a model sees and returns}
\label{app:interface}
Each tick, a model receives one text observation and replies with a JSON array of orders. Figure~\ref{fig:environment-overview}(b) shows part of one observation; the complete text follows. It is what Haiku, playing P2, receives at tick~100 of a \href{https://textcraft-rts.com/\#game/1e6fdcd73ebf4f093f64}{2v2 game on Simple Map Control}: its resources, structures, and units, the enemies in view (\texttt{LIVE INTEL}), the last known positions of those out of view (\texttt{MEMORY}), alerts, the explored terrain, and the last ten ticks of team chat. On the previous tick, Haiku had replied with one order and one message to its teammate:
\begingroup\inprompttrue
\begin{Verbatim}[fontfamily=\promptfontfamily,fontsize=\tiny,breaklines=true,breakanywhere=true,breakautoindent=false,breakindent=0pt,breaksymbolleft={},tabsize=2]
[{"action":"move",
  "unit_ids":["038dc085","1a3124cb","1d008220","2a7512fc","441bb07a","a8796426","b166d40e","e86f540b","fdb1ca75"],
  "target":[19,12]},
 {"action":"team_message",
  "content":"Extractor destroyed. Regrouping 9 soldiers at (19,12). Ready to push P4 CC (20,13) when you arrive—coordinate assault."}]
\end{Verbatim}
\endgroup
\noindent Their effect is visible in the observation, as \texttt{moving to (19, 12)} on the nine soldiers and as the last line of the chat. The system prompt, which states the rules above and the order format, is in the repository.

\begingroup
\setlength{\columnsep}{8pt}
\begin{multicols}{2}
\begingroup\inprompttrue
\begin{Verbatim}[fontfamily=\promptfontfamily,fontsize=\tiny,breaklines=true,breakanywhere=true,breakautoindent=false,breakindent=0pt,breaksymbolleft={},tabsize=2]
=== TICK 100 | Resources: 17 | Income: ~0.0/tick ===

RESOURCES:
  6074f761 at (20, 13) reserve 600/600
  8050b06d at (20, 18) reserve 336/600
  195d9832 at (3, 29) reserve 0/800

YOUR STRUCTURES:
  P2:
    barracks:
      0885f4ba at (2, 26) [100% HP]
    command_center:
      0a5941db at (3, 28) [100% HP]
    extractor:
      44e7b126 at (3, 29) [100% HP] 0 workers, ~1 dist to CC

YOUR UNITS:
  P2:
    soldier:
      038dc085 at (18, 17) [100% HP] cd1 moving to (19, 12)
      1a3124cb at (18, 16) [100% HP] cd1 moving to (19, 12)
      1d008220 at (20, 15) [100% HP] cd1 moving to (19, 12)
      2a7512fc at (23, 15) [100% HP] cd1 moving to (19, 12)
      441bb07a at (21, 15) [100% HP] moving to (19, 12)
      a8796426 at (19, 14) [92% HP] cd1 moving to (19, 12)
      b166d40e at (17, 19) [100% HP] cd1 moving to (19, 12)
      e86f540b at (17, 17) [100% HP] cd1 moving to (19, 12)
      fdb1ca75 at (18, 14) [62% HP] moving to (19, 12)
    worker:
      03c14a74 at (4, 27) [100% HP] moving to (3, 28)
      074e6171 at (2, 27) [100% HP] moving to (3, 28)
      2128c250 at (1, 29) [100% HP] moving to (3, 28)
      3b6fe1f4 at (3, 30) [100% HP] moving to (3, 28)
      517114cb at (2, 29) [100% HP] moving to (3, 28)
      82e3e4b2 at (2, 28) [100% HP] moving to (3, 28)
      8629d801 at (2, 25) [100% HP] moving to (3, 28)
      90065e40 at (3, 27) [100% HP] moving to (3, 28)
      ba9fd56e at (5, 28) [100% HP] moving to (3, 28)
      c3ce48e1 at (4, 29) [100% HP] moving to (3, 28)
      ff52cf68 at (1, 28) [100% HP] moving to (3, 28)

LIVE INTEL:
  P1:
    Units:
      soldier:
        604c168d at (22, 12) [100% HP]
        d9c54382 at (22, 11) [100% HP]
        db97b85d at (21, 11) [100% HP]
  P3:
    Units:
      scout:
        bc4d5982 at (22, 15) [100% HP]
  P4:
    Units:
      scout:
        99099189 at (18, 15) [100% HP]
        c4d23f79 at (22, 18) [100% HP]
        c756de09 at (19, 15) [100% HP]
        fc511069 at (19, 18) [100% HP]
      soldier:
        dce3cc50 at (20, 16) [100% HP]
      worker:
        1589cc66 at (24, 16) [100% HP]

MEMORY:
  Tick 89:
    resource at (11, 13) reserve 600/600
  Tick 88:
    resource at (11, 18) reserve 600/600
  P1:
    Tick 91:
      soldier at (14, 12) (17, 10) (18, 10) (19, 10)
  P3:
    Tick 91:
      scout at (22, 9)
      worker at (15, 9)
    Tick 90:
      worker at (13, 8)
    Tick 88:
      worker at (16, 7)
    Tick 80:
      worker at (25, 5)
  P4:
    Tick 94:
      worker at (25, 23) (29, 23)

ALERTS:
  ENEMY_SPOTTED:
    Enemy scout spotted at (22, 18)

UNKNOWN:
  49% of map unscouted
\end{Verbatim}
\endgroup
\begingroup\inprompttrue
\begin{Verbatim}[fontfamily=\promptfontfamily,fontsize=\tiny,breaklines=true,breakanywhere=true,breakautoindent=false,breakindent=0pt,breaksymbolleft={},tabsize=2]
MINIMAP:
????????????????????????????????
????????????????????????????????
????????????????????????????????
????????????????????????????????
????????????????????????????????
??????????????????......./??????
??????????????.........../??????
??????????????............??????
????????????..............??????
????????????...............?????
???????????................?????
???????????................?????
?????????...................????
????????...r........r.......????
???????.......................??
??????........................??
?????.........................??
????...........................?
???........r........r..........?
??.............................?
?..............................?
#..............................?
#..............................?
#...................???........?
#...................????????????
#//////......?.....?????????????
#^^^^^/......?.....?????????????
#^^^^^/....?????????????????????
#^^^^^/....?????????????????????
#^^R^^/...??????????????????????
#^^^^^/...??????????????????????
#########???????????????????????
\end{Verbatim}
\endgroup
\end{multicols}
\begingroup\inprompttrue
\begin{Verbatim}[fontfamily=\promptfontfamily,fontsize=\tiny,breaklines=true,breakanywhere=true,breakautoindent=false,breakindent=0pt,breaksymbolleft={},tabsize=2]
---
TEAM CHAT:
TICK 91:
P1: Cancelling for extractor at (20,13). Moving soldiers to (14,12) to avoid FF. Please hold fire.
P2 (self): Redirecting 10 soldiers toward P4 workers at (20-21,17). Friendly fire resolved. Ready to coordinate assault when your 5 arrive at (19,12). Target P4 base.
TICK 92:
P1: Canceling structures for extractor at (11,13). My 5 soldiers moving to (19,12) to join you.
P2 (self): Attacking P4 workers at (20,17-19). Your 5 soldiers hold at (14,12)—avoid (19-21,17) area. Once you have income, produce soldiers and push their CC together. I'll weaken their army first.
TICK 93:
P1: Cancelling barracks+depleted extractor. Building extractor at (20,13). Moving soldiers to (14,12). Will produce once income flows.
P2 (self): Eliminating P4 workers at (20-22,17). Hold at (14,12); my soldiers will regroup once workers are dead to push their CC together.
TICK 94:
P1: Cancelling barracks+depleted extractor; building extractor at (11,13). Soldiers holding at (14,12). Will produce soldiers once income flows.
P2 (self): Eliminating P4 workers near (20-23,17). Hold at (14,12)—will regroup 10 soldiers to push P4 CC at (20,13) once workers cleared.
TICK 95:
P1: Moving 5 soldiers to (19,12) to join push. No income—can't build extractor. Let's attack together.
P2 (self): Attacking P4 extractor at (20,18) to cripple economy. Workers harvesting. Ready to push CC once extractor destroyed.
TICK 96:
P1: Attacking P3 workers near (19,12). Scouting (20,13) for extractor — need resources to build it.
P2 (self): Destroying P4 extractor at (20,18) now—10 soldiers focused. Join at (19,12) and we push their CC together.
TICK 97:
P1: Cancelling depleted extractor. Soldiers attacking P3 workers near (20,11) then joining you at (19,12). Sending worker to build extractor on (20,13) after we clear.
P2 (self): Finishing P4 extractor at (20,18)—55% HP. Push their CC at (20,13) once destroyed. Your 5 soldiers heading to (19,12)?
TICK 98:
P1: Cancelling depleted extractor. Soldiers finishing worker near (19,13); will join you at (19,12) after. Let's push their CC together.
P2 (self): Finishing P4 extractor this tick. 9 soldiers focused fire. Regroup at (19,12) after—ready to push their CC together.
TICK 99:
P1: Cancelling structures to build extractor at (20,13). Soldiers attacking P3 worker at (21,12) then regrouping at (19,12)
P2 (self): Finishing P4 extractor (1% HP). Regrouping 9 soldiers at (19,12) to push their CC at (20,13) together. Coordinate assault once you arrive.
TICK 100:
P1: Cancel barracks+extractor for funds. 12 workers moving to (19,12). 5 soldiers finishing P3 worker then joining. Build extractor at (20,13) when worker arrives.
P2 (self): Extractor destroyed. Regrouping 9 soldiers at (19,12). Ready to push P4 CC (20,13) when you arrive—coordinate assault.


Issue your orders for this tick as a JSON array. You may optionally include one {"action":"team_message","content":"..."} object to coordinate with your teammate. If no orders are needed, respond with [].
\end{Verbatim}
\endgroup
\endgroup

\FloatBarrier
\section{Leaderboard Experiments}
\label{app:final-results}

Each of the 18 models in Table~\ref{tab:model-lineup} first plays an elimination bracket among its thinking levels, and then a match between the winning level with and without a scratchpad. The selected configuration enters the Swiss tournament. In all three stages a matchup is ten games: one on each of the five 1v1 maps from each seat, with a 250-tick limit.

\paragraph{Output budget.}
Every call may generate up to 10,240 tokens. If a response reaches this cap before its order array is complete, the scaffold asks once more for the orders alone, with up to 4,096 further tokens. Thinking levels that reach the cap in more than 5\% of calls are not tested. In the tournament the cap is reached in 712 of 201,622 turns (0.35\%), and in 40 turns for the five top-ranked models combined.

\begin{table}[H]
    \centering
    \small
    \setlength{\tabcolsep}{5pt}
    \renewcommand{\arraystretch}{1.08}

    \caption{The 18 models, the thinking levels tested for each, and the configuration selected for the tournament. Scratchpad: \scratchpadon{} selected on; \scratchpadoff{} selected off. Colors identify model families.}
    \label{tab:model-lineup}
    \label{tab:selected-configs}

    \begin{tabularx}{\textwidth}{@{} l l >{\raggedright\arraybackslash}X l c @{}}
        \toprule
        \textbf{Model} & \textbf{Weights} & \textbf{Thinking levels tested} & \textbf{Selected} & \textbf{Scratchpad} \\
        \midrule
        \multicolumn{5}{@{}l@{}}{\textbf{Large}} \\
        \addlinespace[2pt]
        \modelname{familyGPT}{GPT-6 Astra} & Proprietary & Low, medium, high, xhigh & xhigh & \scratchpadoff \\
        \modelname{familyGPT}{GPT-5.6 Sol} & Proprietary & None, low, high, xhigh & high & \scratchpadoff \\
        \modelname{familyClaude}{Claude Opus 5} & Proprietary & Disabled; adaptive low, high & high (adaptive) & \scratchpadoff \\
        \modelname{familyClaude}{Claude Fable 5.1} & Proprietary & Adaptive low, medium, high & high (adaptive) & \scratchpadoff \\
        \modelname{familyGrok}{Grok 4.6} & Proprietary & Low, medium & medium & \scratchpadoff \\
        \modelname{familyQwen}{Qwen3.8-2.4T-A95B} & Open-weight & Low, medium & medium & \scratchpadon \\
        \modelname{familyKimi}{Kimi K3} & Open-weight & Low, high & high & \scratchpadon \\
        \modelname{familyGemini}{Gemini 3.1 Pro} & Proprietary & Low, high & low & \scratchpadoff \\
        \modelname{familyDeepSeek}{DeepSeek V4 Pro} & Open-weight & Off; low, high & low & \scratchpadoff \\
        \midrule
        \multicolumn{5}{@{}l@{}}{\textbf{Medium}} \\
        \addlinespace[2pt]
        \modelname{familyGemini}{Gemini 3.7 Flash} & Proprietary & Low, high & high & \scratchpadoff \\
        \modelname{familyClaude}{Claude Sonnet 5} & Proprietary & Disabled; adaptive high & high (adaptive) & \scratchpadoff \\
        \modelname{familyDeepSeek}{DeepSeek V4 Flash} & Open-weight & Off; low, high & high & \scratchpadoff \\
        \modelname{familyGLM}{GLM 5.3} & Open-weight & Low, high & high & \scratchpadoff \\
        \modelname{familyGPT}{GPT-5.6 Terra} & Proprietary & None, low, high, xhigh & xhigh & \scratchpadoff \\
        \midrule
        \multicolumn{5}{@{}l@{}}{\textbf{Small}} \\
        \addlinespace[2pt]
        \modelname{familyGPT}{GPT-5.6 Luna} & Proprietary & None, low, high, xhigh & xhigh & \scratchpadoff \\
        \modelname{familyClaude}{Claude Haiku 4.5} & Proprietary & Off; on with a 1,024-token budget & off & \scratchpadoff \\
        \modelname{familyQwen}{Qwen3.8-27B} & Open-weight & Disabled, low, medium & medium & \scratchpadoff \\
        \modelname{familyGLM}{GLM 5.3 Flash} & Open-weight & Low, high & high & \scratchpadon \\
        \bottomrule
    \end{tabularx}
\end{table}

\subsection{Thinking-Variant Selection}
\label{app:thinking-results}

\begin{figure}[!htbp]
\centering
\includegraphics[width=\linewidth,height=.80\textheight,keepaspectratio]{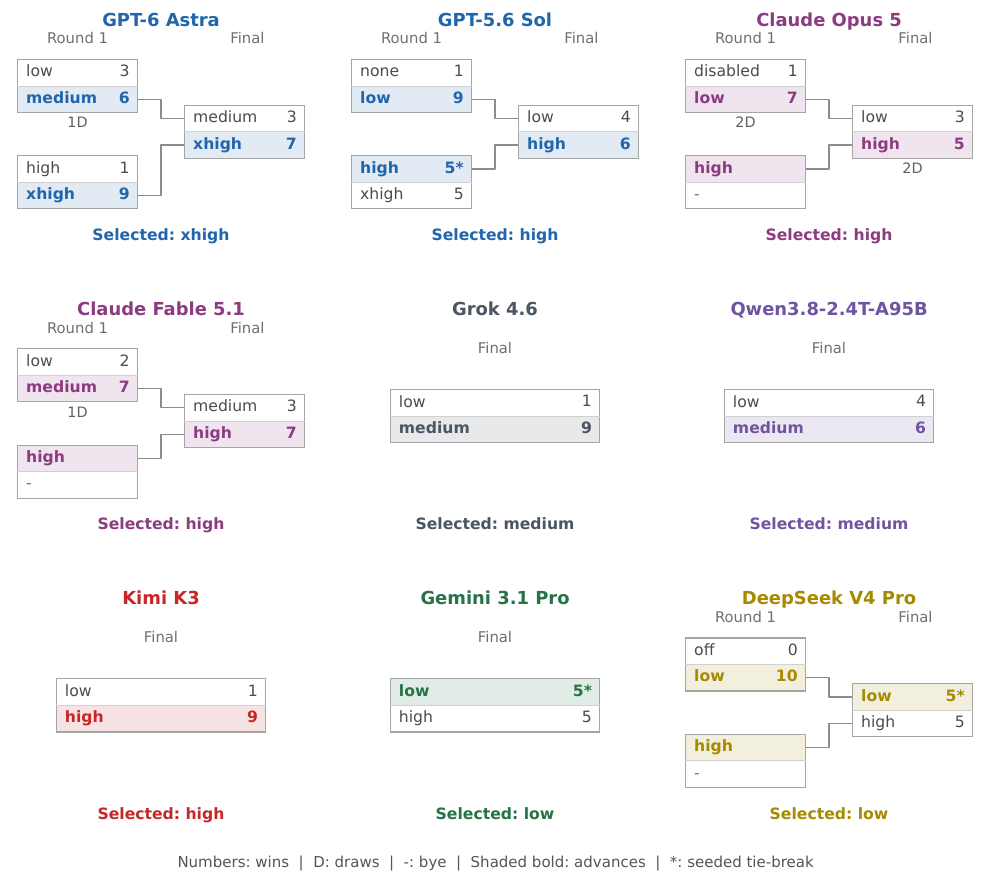}
\caption{Thinking-Variant Selection brackets for the first nine models. Each match is ten games; numbers are wins, and D counts draws. Shaded rows advance; an asterisk marks advancement on the tie-break, and \texttt{-} a bye.}
\label{fig:thinking}
\end{figure}

\begin{figure}[!htbp]
\centering
\includegraphics[width=\linewidth,height=.80\textheight,keepaspectratio]{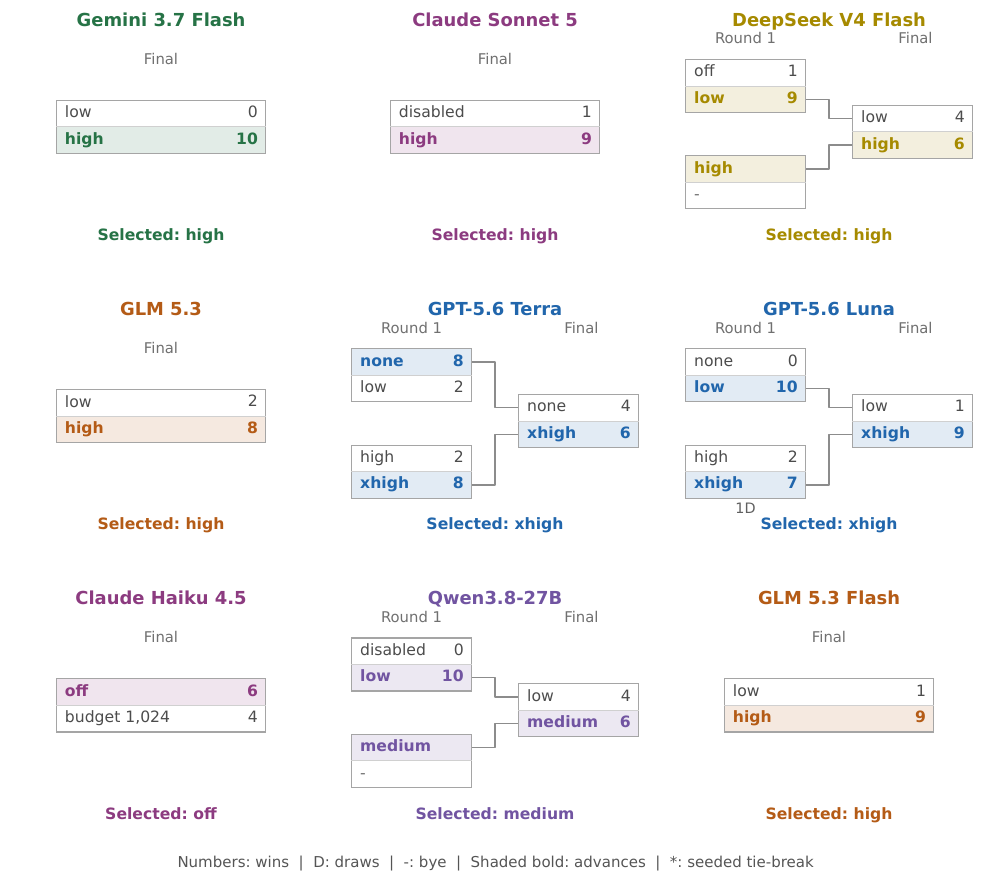}
\caption{Thinking-Variant Selection brackets for the remaining nine models.}
\label{fig:thinking-rest}
\end{figure}

The thinking levels of each model meet in a single-elimination bracket that starts from the lowest level: 49 entrants, 31 matchups, and 310 games, all without a scratchpad. On equal wins the lower level advances. The highest tested level wins 14 of the 18 brackets (Figures~\ref{fig:thinking} and~\ref{fig:thinking-rest}). The exceptions are GPT-5.6 Sol (high, after a 5--5 tie), Gemini 3.1 Pro (low, after a 5--5 tie), DeepSeek V4 Pro (low, after a tied 5--5 final against high), and Haiku 4.5 (thinking off, 6--4 against its 1,024-token thinking variant).

\FloatBarrier
\subsection{Scratchpad Selection}
\label{app:scratchpad-results}

When enabled, the scratchpad is a private text buffer of at most \ScratchpadCap{} characters. The model rewrites it by adding an \texttt{update\_scratchpad} object to its order array, and the buffer is shown to the model at the start of every later tick. It is never sent to the engine or to other players.

\begin{figure}[!htbp]
\centering
\includegraphics[width=\linewidth]{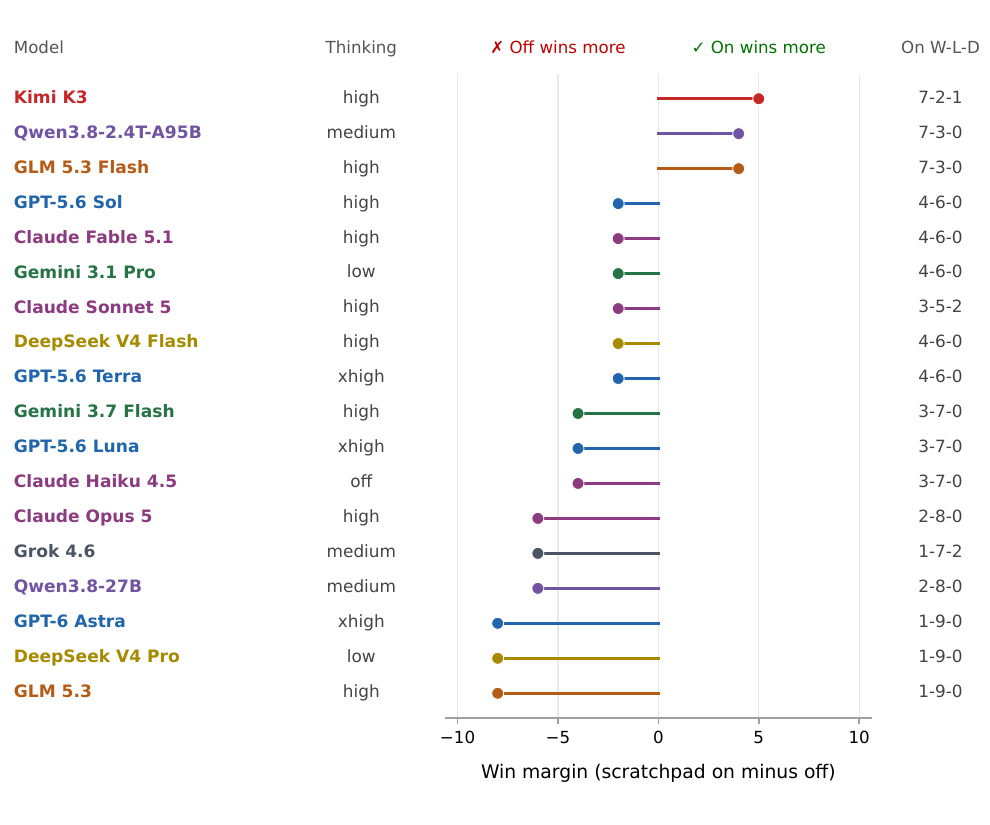}
\caption{Scratchpad Selection: wins with the scratchpad minus wins without it, ten games per model. The left column gives the thinking level used on both sides, and the right column the scratchpad side's wins--losses--draws.}
\label{fig:scratchpad}
\end{figure}

Each model plays ten games with the scratchpad against itself without one, at its selected thinking level (180 games). The scratchpad wins only for Qwen3.8-2.4T-A95B (7--3--0), Kimi K3 (7--2--1), and GLM 5.3 Flash (7--3--0). Over all 180 games, the scratchpad side wins 61, loses 114, and draws five (Figure~\ref{fig:scratchpad}).

\FloatBarrier
\subsection{Swiss Leaderboard}
\label{app:swiss-results}

\begin{figure}[!htbp]
\centering
\includegraphics[width=\linewidth]{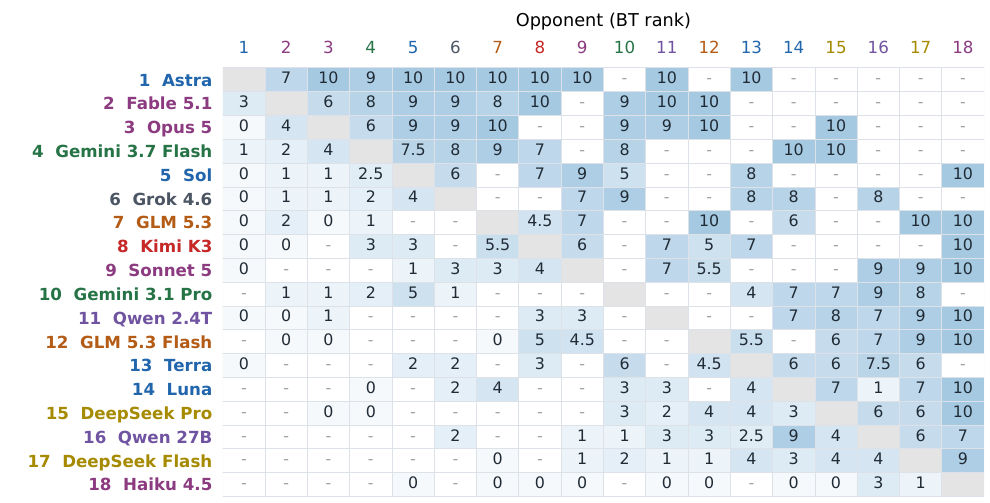}
\caption{Pairwise scores after ten Swiss rounds. Each cell is the row model's score against the column model over ten games, with one point per win and half per draw; \texttt{-} marks pairs that did not meet.}
\label{fig:swiss-pairwise}
\end{figure}

The selected configurations play ten Swiss rounds, 900 games. Five games are draws, and the two starting positions win 444 and 451 games. Table~\ref{tab:leaderboard} gives the final standings, Figure~\ref{fig:swiss-pairwise} the result of every pairing, and Table~\ref{tab:leaderboard-metrics} per-game averages.

\paragraph{Rating estimation.}
Bradley--Terry (BT) ratings count a draw as half a win and use a symmetric prior of 0.01 games per pair of players. Abilities $a_i$ are normalized to a geometric mean of one and reported as $400\log_{10}a_i$, so ratings are centered on zero. Intervals are 95\% percentile intervals from 10,000 bootstrap resamples of the games within each pairing.

\begin{table}[!htbp]
\centering\small
\setlength{\tabcolsep}{5pt}
\caption{Final Swiss Leaderboard, 100 games per model. P1 and P2 are win rates (\%) from each starting position, 50 games each.}
\label{tab:leaderboard}
\begin{tabular}{@{}rlrcrrr@{}}
\toprule
Rank & Model & BT & 95\% interval & W--L--D & P1 & P2 \\
\midrule
\modelname{familyGPT}{1} & \modelname{familyGPT}{GPT-6 Astra} & $796.7$ & $[681,\,1053]$ & 96--4--0 & 94 & 98 \\
\modelname{familyClaude}{2} & \modelname{familyClaude}{Claude Fable 5.1} & $517.6$ & $[413,\,674]$ & 82--18--0 & 78 & 86 \\
\modelname{familyClaude}{3} & \modelname{familyClaude}{Claude Opus 5} & $424.8$ & $[342,\,548]$ & 76--24--0 & 72 & 80 \\
\modelname{familyGemini}{4} & \modelname{familyGemini}{Gemini 3.7 Flash} & $310.6$ & $[222,\,425]$ & 66--33--1 & 64 & 68 \\
\modelname{familyGPT}{5} & \modelname{familyGPT}{GPT-5.6 Sol} & $115.7$ & $[33,\,206]$ & 49--50--1 & 48 & 50 \\
\modelname{familyGrok}{6} & \modelname{familyGrok}{Grok 4.6} & $84.7$ & $[{-}1,\,179]$ & 48--52--0 & 44 & 52 \\
\modelname{familyGLM}{7} & \modelname{familyGLM}{GLM 5.3} & $53.8$ & $[{-}32,\,143]$ & 50--49--1 & 50 & 50 \\
\modelname{familyKimi}{8} & \modelname{familyKimi}{Kimi K3} & $14.1$ & $[{-}73,\,101]$ & 46--53--1 & 54 & 38 \\
\modelname{familyClaude}{9} & \modelname{familyClaude}{Claude Sonnet 5} & $-51.2$ & $[{-}128,\,24]$ & 51--48--1 & 50 & 52 \\
\modelname{familyGemini}{10} & \modelname{familyGemini}{Gemini 3.1 Pro} & $-58.6$ & $[{-}154,\,34]$ & 45--55--0 & 48 & 42 \\
\modelname{familyQwen}{11} & \modelname{familyQwen}{Qwen3.8-2.4T-A95B} & $-95.0$ & $[{-}186,\,{-}6]$ & 48--52--0 & 46 & 50 \\
\modelname{familyGLM}{12} & \modelname{familyGLM}{GLM 5.3 Flash} & $-126.0$ & $[{-}210,\,{-}52]$ & 46--52--2 & 50 & 42 \\
\modelname{familyGPT}{13} & \modelname{familyGPT}{GPT-5.6 Terra} & $-141.2$ & $[{-}229,\,{-}65]$ & 42--56--2 & 46 & 38 \\
\modelname{familyGPT}{14} & \modelname{familyGPT}{GPT-5.6 Luna} & $-223.8$ & $[{-}316,\,{-}147]$ & 41--59--0 & 46 & 36 \\
\modelname{familyDeepSeek}{15} & \modelname{familyDeepSeek}{DeepSeek V4 Pro} & $-257.2$ & $[{-}352,\,{-}182]$ & 38--62--0 & 40 & 36 \\
\modelname{familyQwen}{16} & \modelname{familyQwen}{Qwen3.8-27B} & $-277.7$ & $[{-}366,\,{-}205]$ & 38--61--1 & 28 & 48 \\
\modelname{familyDeepSeek}{17} & \modelname{familyDeepSeek}{DeepSeek V4 Flash} & $-362.6$ & $[{-}462,\,{-}290]$ & 29--71--0 & 26 & 32 \\
\modelname{familyClaude}{18} & \modelname{familyClaude}{Claude Haiku 4.5} & $-724.8$ & $[{-}969,\,{-}617]$ & 4--96--0 & 4 & 4 \\
\bottomrule
\end{tabular}
\end{table}

\begin{table}[!htbp]
\centering\small
\setlength{\tabcolsep}{5pt}
\caption{Per-game averages in the tournament. Economy is the resources added to a player's balance per tick; valid orders are counted per unit or structure ordered. Cost per tick in Figure~\ref{fig:rating-cost} is a model's total API cost divided by its ticks played.}
\label{tab:leaderboard-metrics}
\begin{tabular}{@{}rlrrr@{}}
\toprule
Rank & Model & Economy/tick & Valid orders/tick & Cost (USD/game) \\
\midrule
\modelname{familyGPT}{1} & \modelname{familyGPT}{GPT-6 Astra} & 10.93 & 2.64 & 6.89 \\
\modelname{familyClaude}{2} & \modelname{familyClaude}{Claude Fable 5.1} & 10.47 & 2.67 & 6.93 \\
\modelname{familyClaude}{3} & \modelname{familyClaude}{Claude Opus 5} & 10.02 & 2.99 & 3.76 \\
\modelname{familyGemini}{4} & \modelname{familyGemini}{Gemini 3.7 Flash} & 9.99 & 2.30 & 1.23 \\
\modelname{familyGPT}{5} & \modelname{familyGPT}{GPT-5.6 Sol} & 9.39 & 2.08 & 3.63 \\
\modelname{familyGrok}{6} & \modelname{familyGrok}{Grok 4.6} & 8.94 & 4.24 & 3.60 \\
\modelname{familyGLM}{7} & \modelname{familyGLM}{GLM 5.3} & 8.86 & 2.34 & 1.08 \\
\modelname{familyKimi}{8} & \modelname{familyKimi}{Kimi K3} & 9.25 & 1.89 & 4.89 \\
\modelname{familyClaude}{9} & \modelname{familyClaude}{Claude Sonnet 5} & 8.85 & 2.11 & 3.12 \\
\modelname{familyGemini}{10} & \modelname{familyGemini}{Gemini 3.1 Pro} & 8.78 & 2.05 & 1.04 \\
\modelname{familyQwen}{11} & \modelname{familyQwen}{Qwen3.8-2.4T-A95B} & 8.48 & 1.74 & 2.13 \\
\modelname{familyGLM}{12} & \modelname{familyGLM}{GLM 5.3 Flash} & 8.68 & 1.98 & 0.23 \\
\modelname{familyGPT}{13} & \modelname{familyGPT}{GPT-5.6 Terra} & 8.85 & 1.79 & 2.44 \\
\modelname{familyGPT}{14} & \modelname{familyGPT}{GPT-5.6 Luna} & 8.06 & 1.49 & 0.25 \\
\modelname{familyDeepSeek}{15} & \modelname{familyDeepSeek}{DeepSeek V4 Pro} & 8.98 & 2.47 & 1.65 \\
\modelname{familyQwen}{16} & \modelname{familyQwen}{Qwen3.8-27B} & 8.95 & 1.75 & 0.81 \\
\modelname{familyDeepSeek}{17} & \modelname{familyDeepSeek}{DeepSeek V4 Flash} & 8.27 & 3.06 & 0.28 \\
\modelname{familyClaude}{18} & \modelname{familyClaude}{Claude Haiku 4.5} & 6.79 & 2.16 & 0.46 \\
\bottomrule
\end{tabular}
\end{table}

\FloatBarrier
\subsection{Cost and data volume}
\label{app:data-volume}
The experiments comprise 4,670 games, 1,008,304 model turns, and 1,700,339 submitted orders, at an API cost of \$15,734 (Table~\ref{tab:final-costs}). Skill learning (Appendix~\ref{app:skills}) accounts for 3,208 of the games and \$7,043 of the cost, \$378 of it for the 136 teacher sessions. The 1,462 games of the three leaderboard experiments and the 2v2 evaluation of Appendix~\ref{app:team-tsne} are released with their observations, orders, replay, and outcome, and with team messages, scratchpads, and reasoning traces where they exist, and can be browsed at \url{https://textcraft-rts.com/}.

\begin{table}[!htbp]
\centering\small
\setlength{\tabcolsep}{5pt}
\caption{Size and API cost of the experiments. M is $10^6$ tokens. Skill learning covers the games reported in Appendix~\ref{app:skills}; its tokens and cost include the teacher sessions.}
\label{tab:final-costs}
\label{tab:experiment-data-volume}
\begin{tabular}{@{}lrrrrrr@{}}
\toprule
 & \shortstack[r]{Thinking\\variants} & Scratchpad & \shortstack[r]{Swiss\\leaderboard} & \shortstack[r]{Team\\chat} & \shortstack[r]{Skill\\learning} & All \\
\midrule
Completed games & 310 & 180 & 900 & 72 & 3,208 & 4,670 \\
Game ticks & 36,320 & 22,713 & 100,811 & 9,919 & 356,463 & 526,226 \\
Model turns & 72,640 & 45,426 & 201,622 & 37,531 & 651,085 & 1,008,304 \\
Orders submitted & 131,299 & 82,348 & 374,996 & 70,500 & 1,041,196 & 1,700,339 \\
Teacher sessions & -- & -- & -- & -- & 136 & 136 \\
\midrule
Input tokens (M) & 442 & 290 & 1,255 & 291 & 4,749 & 7,027 \\
Output tokens (M) & 75 & 84 & 365 & 53 & 569 & 1,146 \\
API cost (USD) & 1,746 & 1,125 & 4,442 & 1,378 & 7,043 & 15,734 \\
\bottomrule
\end{tabular}
\end{table}

\FloatBarrier
\section{Validity Checks: Reading the Interface and Engine Exploits}
\label{app:validity}

\subsection{Observation-reading quiz}
\label{app:obs-quiz}
We test whether models can read the text observation, separately from how well they play. We sampled 40 observations from recorded 1v1 games on the five maps, half from each seat and spread over the early, middle, and late game, and generated five questions for each, 200 in total. \emph{Lookup} questions (106) ask for a fact stated in the observation, such as ``How many soldier units do you currently have?''. \emph{Inference} questions (94) need one step of reasoning over the map, such as ``Without moving, which of your units can attack the enemy unit X this tick?''. Gold answers come from the engine's recorded state, and a question is kept only if an independent oracle that sees the observation text alone derives the same answer. Each model receives the game rules and one observation per call, with high thinking, and answers are scored by exact match.

\begin{table}[!htbp]
\centering\small
\setlength{\tabcolsep}{6pt}
\caption{\textbf{Models read the observation almost equally well and play very differently.} Correct answers on the observation-reading quiz (106 lookup and 94 inference questions) and wins in the 100 tournament games of Table~\ref{tab:leaderboard}. Rows follow tournament rank.}
\label{tab:obs-quiz}
\begin{tabular}{@{}lrrrr@{}}
\toprule
Model & Lookup / 106 & Inference / 94 & Overall \% & Tournament wins \\
\midrule
\modelname{familyGPT}{GPT-6 Astra} & 106 & 94 & 100 & 96 \\
\modelname{familyClaude}{Claude Fable 5.1} & 106 & 94 & 100 & 82 \\
\modelname{familyClaude}{Claude Opus 5} & 105 & 93 & 99 & 76 \\
\modelname{familyGemini}{Gemini 3.7 Flash} & 106 & 94 & 100 & 66 \\
\modelname{familyGPT}{GPT-5.6 Sol} & 104 & 94 & 99 & 49 \\
\modelname{familyGemini}{Gemini 3.1 Pro} & 106 & 94 & 100 & 45 \\
\modelname{familyGPT}{GPT-5.6 Terra} & 103 & 89 & 96 & 42 \\
\modelname{familyGPT}{GPT-5.6 Luna} & 101 & 91 & 96 & 41 \\
\modelname{familyClaude}{Claude Haiku 4.5} & 106 & 88 & 97 & 4 \\
\bottomrule
\end{tabular}
\end{table}

All nine models answer between 96\% and 100\% of the questions (Table~\ref{tab:obs-quiz}). Quiz scores do not follow playing strength: Gemini~3.1 Pro answers every question and wins 45 tournament games, while Opus misses two and wins 76. Haiku, which wins four games, reads every lookup question correctly. Differences in play therefore come from what models do with the game state, and the text interface is not what limits them.

\subsection{Replay audit}
\label{app:replay-audit}
We audited the tournament for wins that come from exploiting the engine. Two frontier models, GPT-6 Astra and Claude Fable~5.2, independently reviewed the 270 games played by the three top-ranked models. Besides reading the games, they wrote and ran scripted checks that compare orders and attacks with the recorded game state, so the result does not rest on a model's judgment alone; these checks ran over all 900 tournament games.

In the 270 games, we found no violation of the rules for resources, hit points, terrain, attack range, movement speed, or rate of fire. None of the three top models ordered an attack on a target outside its vision (0 of 19,925 attack orders), and neither review found a model relying on unintended engine behavior. This fits how models interact with the game: they act only through JSON orders, cannot see the engine's code, and carry no memory from one tournament game to the next. All replays and traces are public, so the check can be repeated.

\FloatBarrier
\section{Behavioral Profiles in the 1v1 Tournament}
\label{app:swiss-tsne}

Each of the 900 tournament games gives two points, one per player: 1,800 player-games, 100 per model. We describe every player-game by what happened in it (its economy, production, army composition, combat, and positioning, and how these develop over time) and group the player-games into ten \emph{profiles}, A1--A10 (Figure~\ref{fig:swiss-strategy-performance}a,b). A profile describes how a game unfolded, including how it ended. Appendix~\ref{app:swiss-clustering} gives the procedure.

\subsection{The ten profiles}
\label{app:swiss-profiles}

\begin{table}[!htbp]
\centering\footnotesize
\caption{The ten profiles. Mining is resources mined per tick, army value is the cost of the living army weighted by its remaining HP, and forward is the share of the army beyond the midpoint between the two bases. Each pair gives the mean over the two tick windows shown.}
\label{tab:swiss-temporal-profiles}
\setlength{\tabcolsep}{3pt}
\renewcommand{\arraystretch}{1.12}
\begin{tabular}{@{}c>{\centering\arraybackslash}p{0.255\linewidth}cr@{\,$\to$\,}lr@{\,$\to$\,}lr@{\,$\to$\,}lc@{}}
\toprule Profile & Name & \shortstack{Player-\\games} & \multicolumn{2}{c}{Mining} & \multicolumn{2}{c}{Army value} & \multicolumn{2}{c}{Forward (\%)} & Win (\%) \\
 & & & 21--40 & 61--80 & 41--60 & 61--80 & 41--60 & 61--80 & \\
\midrule
A1 & Order errors and delayed army & 97 & 10.9 & 7.6 & 29 & 44 & 3 & 2 & 2.1 \\
A2 & Early mining collapse & 231 & 16.0 & 4.4 & 94 & 45 & 11 & 6 & 4.8 \\
A3 & Expansion with mining decline & 161 & 16.7 & 4.5 & 133 & 85 & 17 & 11 & 10.6 \\
A4 & Striker-heavy army decline & 192 & 17.3 & 15.2 & 125 & 88 & 12 & 5 & 2.6 \\
A5 & Soldier attrition & 184 & 17.0 & 17.9 & 154 & 165 & 21 & 15 & 17.4 \\
A6 & Technology investment & 53 & 17.2 & 18.4 & 166 & 197 & 22 & 22 & 49.1 \\
A7 & Expansion and army persistence & 284 & 17.0 & 5.4 & 185 & 222 & 32 & 62 & 95.8 \\
A8 & Sustained soldier buildup & 194 & 16.9 & 18.5 & 221 & 369 & 27 & 52 & 90.2 \\
A9 & Gradual mixed-army buildup & 169 & 18.3 & 18.4 & 186 & 263 & 21 & 33 & 76.9 \\
A10 & Marauder-led opening & 235 & 18.0 & 15.8 & 246 & 400 & 55 & 79 & 95.7 \\
\bottomrule
\end{tabular}
\end{table}

Table~\ref{tab:swiss-temporal-profiles} gives the size, development, and win rate of each profile. The statistics that mark each one follow, as shares of its player-games. A \emph{sustained forward army} is at least three military units, at least half of them beyond the midpoint, for ten consecutive ticks.

\begin{description}[leftmargin=1.2em,itemsep=2pt,topsep=4pt]
\item[A1, order errors and delayed army.] In 94.8\%, at least a fifth of the ticks carry an \texttt{order\_failed} alert. The median first military unit appears at tick 51, and 4.1\% develop a sustained forward army.
\item[A2, early mining collapse.] In 58.0\%, mining stops for ten ticks or more once the starting deposit is depleted. 5.6\% develop a sustained forward army.
\item[A3, expansion with mining decline.] 88.2\% hold two extractors at once, yet 89.4\% still have a ten-tick mining gap after the starting deposit is depleted, and 25.5\% develop a sustained forward army.
\item[A4, striker-heavy army decline.] Production is striker-heavy and the army shrinks. Repair orders on the command center recur in 79.2\%, and 12.5\% develop a sustained forward army.
\item[A5, soldier attrition.] In 67.4\%, at least four military units are completed and at least 75\% of them are soldiers, but the army barely grows. Repair orders on the command center recur in 72.3\%.
\item[A6, technology investment.] An upgrade is researched in 83.0\%, and tanks or enforcers are built in 81.1\%.
\item[A7, expansion and army persistence.] 90.1\% hold two extractors at once, as in A3, but 95.4\% develop a sustained forward army.
\item[A8, sustained soldier buildup.] Production is soldier-dominated in 90.7\%, as in A5, but the army keeps growing, and 98.5\% develop a sustained forward army.
\item[A9, gradual mixed-army buildup.] In 62.1\%, at least two of the three infantry types are completed by tick 50. 88.2\% develop a sustained forward army.
\item[A10, marauder-led opening.] In 75.7\%, at least three military units are completed by tick 50 and at least half of them are marauders. 99.1\% develop a sustained forward army.
\end{description}

\subsection{Profiles by model}
\begin{figure}[!htbp]
\centering
\includegraphics[width=\linewidth]{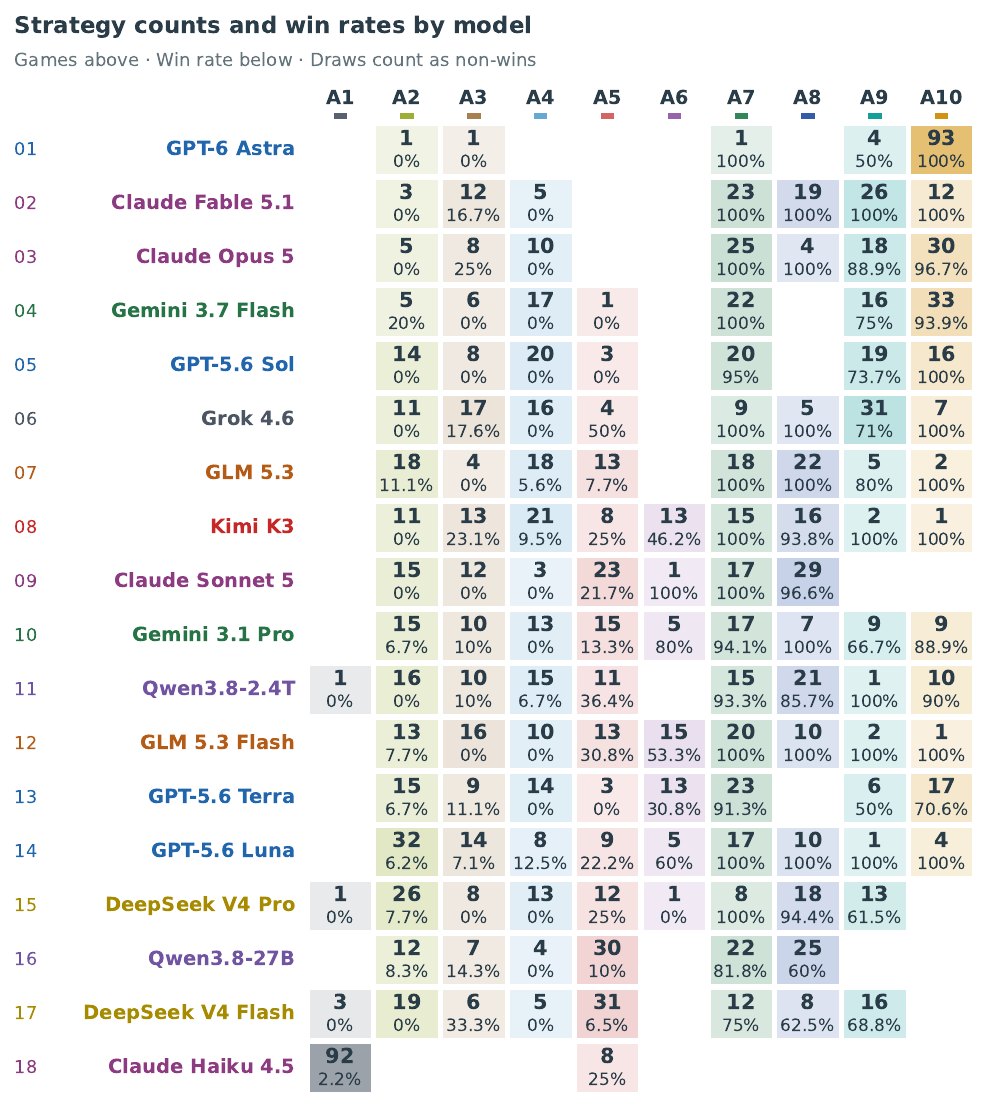}
\caption{\textbf{Profiles by model.} Each cell gives the number of a model's player-games in a profile (top) and their win rate (bottom). Rows follow tournament rank, and colors identify the profiles of Figure~\ref{fig:swiss-strategy-performance}.}
\label{fig:swiss-appendix-models}
\end{figure}

Figure~\ref{fig:swiss-appendix-models} gives, for every model, the number of its 100 player-games in each profile and how many of them it wins.

\subsection{How Astra's opening is beaten}
\label{app:astra-traces}
Astra wins 96 of its 100 games. 93 of them are in A10, the marauder-led opening, and it wins all 93; its other three wins are in A7 (one) and A9 (two). Its four losses start from the same opening, with marauders making up 80--100\% of its military units through tick 50.

\paragraph{Most counters fail.}
Fable produces strikers in all six of its losses to an A10 Astra. In the \href{https://textcraft-rts.com/\#game/33fb0eada7977791ee04}{King of the Hill example} (game 83), it completes seven strikers, but Astra keeps its marauder core, adds soldiers, and holds a forward army from ticks 65--95.

\paragraph{Fable: counter-production from what it observes.}
In its three wins, Fable sees Astra's marauders, names strikers as the counter in its reasoning, and orders them.

In \href{https://textcraft-rts.com/\#game/74dfefff30be2c4c2fa1}{Simple Map Control} (game 66), Fable observes a marauder at tick 37, orders a striker at tick 38, and pulls its soldiers back from the unfavorable engagement. It completes four strikers by tick 50 and seven by tick 58. Astra's forward presence ends at tick 59, and its remaining marauders are eliminated by tick 75. Despite a second extractor at tick 32, it cannot maintain its army or income, and its game ends in A3.

In \href{https://textcraft-rts.com/\#game/835203a37147a13308d6}{Simple} (game 78), Fable orders its first striker at tick 37. Three strikers are completed by tick 45, and four of Astra's first five marauders are dead by tick 55. Astra adds soldiers and later rebuilds its marauder force, while Fable shifts toward marauders of its own. Astra's game ends in A9.

In \href{https://textcraft-rts.com/\#game/6f3e10ac4110d4d358b1}{Simple Mounds} (game 89), Fable completes six strikers by tick 50, and its tick-48 orders send pairs of strikers at individual marauders. Astra starts adding soldiers at tick 49, but all four opening marauders are gone by tick 58. Fable notes the change to soldiers and orders marauders at tick 60 to counter them. Astra later turns to strikers, and its game ends in A9.

\paragraph{Gemini Flash: better exchanges with the same units.}
In the \href{https://textcraft-rts.com/\#game/c21278f56990e2592f0e}{King of the Hill loss} (game 448), both players produce only marauders through tick 50. Gemini Flash's reasoning at ticks 43--44 weighs the positions of the observed marauders, attack range, the downhill damage bonus, and focus fire. Its orders send three marauders at one target and then at the next, and Astra's two marauders die at ticks 43 and 45. In each exchange, two of the attackers fire from high ground onto low ground.

By tick 60, Gemini Flash has killed six marauders and lost one, although it has completed only four against Astra's seven. It uses the surviving force to contest Astra's expansion. Astra never establishes a sustained forward army, has no mining from ticks 73--98, and its game ends in A2. Gemini Flash adds strikers only from tick 71.

\subsection{Scripted bots}
\label{app:swiss-bots}
\begin{figure}[!htbp]
\centering
\includegraphics[width=\linewidth]{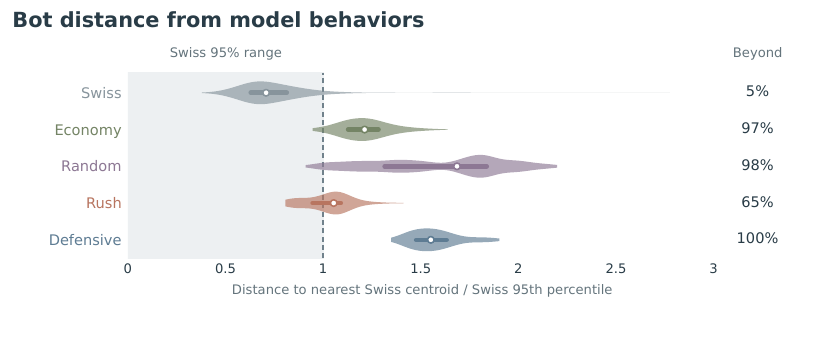}
\caption{\textbf{Scripted bots play unlike the language models.} Distance from each player-game to the nearest profile center, relative to the distance that contains 95\% of tournament player-games (shaded). Dots are medians and bars the middle 50\%; the right column gives the share beyond the threshold.}
\label{fig:swiss-bot-distance}
\end{figure}

We ran 600 games between the four scripted bots on Simple, 300 player-games per bot, and measured how far each lies from the nearest profile center, relative to the distance that contains 95\% of tournament player-games (Figure~\ref{fig:swiss-bot-distance}). 96.7\% of Economy, 97.7\% of Random, 65.3\% of Rush, and 100\% of Defensive player-games lie beyond that distance.

\subsection{How the profiles are computed}
\label{app:swiss-clustering}

\begin{methodbox}{From replays to profiles}
\begin{enumerate}[leftmargin=1.4em,itemsep=2pt,topsep=0pt]
\item \textbf{Points.} One point per player per game: 1,800 player-games.
\item \textbf{Full-game features.} 57 summaries of the whole game in eight groups (Table~\ref{tab:swiss-tsne-features}), computed from replay states and executed events.
\item \textbf{Temporal features.} 14 quantities averaged over seven 20-tick windows, from tick 1 to tick 140: 98 values.
\item \textbf{Missing values.} A ratio whose opportunity never arises, such as the success rate of uphill assaults in a game without one, takes the feature mean, and a binary flag marks it as missing.
\item \textbf{Scaling and weights.} Features are standardized, temporal magnitudes after $\log(1+x)$. Feature groups weigh equally, features within a group weigh equally, and the flags carry 5\% of the weight.
\item \textbf{Combination.} The full-game and temporal blocks are scaled to the same total variance and concatenated: 270 columns.
\item \textbf{Clustering.} $k$-means with $k=10$ on these vectors. Model, opponent, map, seat, outcome, and cost are not inputs.
\item \textbf{Display and names.} t-SNE with perplexity 30 on the same vectors draws Figure~\ref{fig:swiss-strategy-performance}(a). Profiles are numbered from left to right in that plot and named after inspecting their features.
\end{enumerate}
\end{methodbox}

\begin{table}[H]
\centering\footnotesize
\caption{Features by group, with the number in each group. Military units are soldiers, marauders, strikers, tanks, and enforcers.}
\label{tab:swiss-tsne-features}
\renewcommand{\arraystretch}{1.15}
\begin{tabularx}{\linewidth}{@{}l>{\raggedright\arraybackslash}X@{}}
\toprule
Group & Features \\
\midrule
Economy (7) & Mining rate; mean workers, extractors, and bank; idle-worker share; share of the map's remaining resources under own extractors; mining after the starting node is depleted. \\
Production (9) & Mean barracks and their busy share; military production value; military share of unit spending; scout production; idle barracks while resources are available; technology structures; upgrades completed and active. \\
Composition (9) & Share of each of the six non-worker unit types; entropy of the units produced and of the units fielded; share of production that counters the visible enemy army. \\
Combat (7) & Army value weighted by HP; value killed and lost; workers killed; focus fire; repair orders on a damaged command center; attacks on structures while enemy units are near. \\
Position (4) & Share of the map explored; army share beyond the midpoint; isolated units; units holding or without orders. \\
Commands (4) & Orders and attack orders per tick; ticks without a new order; ticks with a failed order. \\
Terrain (11) & Downhill and uphill attack shares; downhill fire received; downhill fire and high-ground positions in defense; frequency of downhill combat; uphill focus fire; uphill assaults that reach the plateau, keep attacking there, hold it, or are lost on the way. \\
Unit matchups (6) & Shares of attacks made, and of attacks received, in favorable, unfavorable, and neutral matchups. \\
\midrule
Temporal (14 per window) & Mining, workers, extractors, and bank; barracks, their busy share, and military production value; soldier, marauder, striker, and advanced-unit shares of production; army size and HP-weighted value; army share beyond the midpoint. \\
\bottomrule
\end{tabularx}
\end{table}

\noindent The exact definition of every feature and all settings are in the repository.

\FloatBarrier
\section{Skill Learning: Protocol and Results}
\label{app:skills}
A teacher model (GPT-5.6 Sol, high reasoning) reads a student's games and writes a \emph{handbook}: a short set of natural-language skills that is added to the student's prompt in later games. Each generation adds training games and produces a new handbook; G$n$ is the handbook after generation $n$, and G0 is the student without one. Unless stated otherwise, games are played on \texttt{simple\_2p}. Single-map evaluations run for at most 200 ticks; the five-map panel of Appendix~\ref{app:skills-regimes} uses the tournament's 250.

\subsection{Stateful versus stateless teachers}
\label{app:skills-teacher}
Haiku trains against four bots (Random, Economy, Rush, and Defensive) for twelve generations of four games, one per bot, with alternating seats. A handbook holds at most eight skills and 8,000 characters. Both teachers can read every training game played so far. The \emph{stateful} teacher also sees its current handbook and the order of the generations. The \emph{stateless} teacher receives the games anonymized and unordered, sees no earlier handbook, and writes each handbook from scratch, with a larger tool-call budget. Each teacher is run once. We evaluate every checkpoint over twenty games against Sonnet~4.6, which is never used in training (this experiment predates the tournament; later evaluations use Sonnet~5).

\begin{figure}[!ht]
\centering
\includegraphics[width=\linewidth]{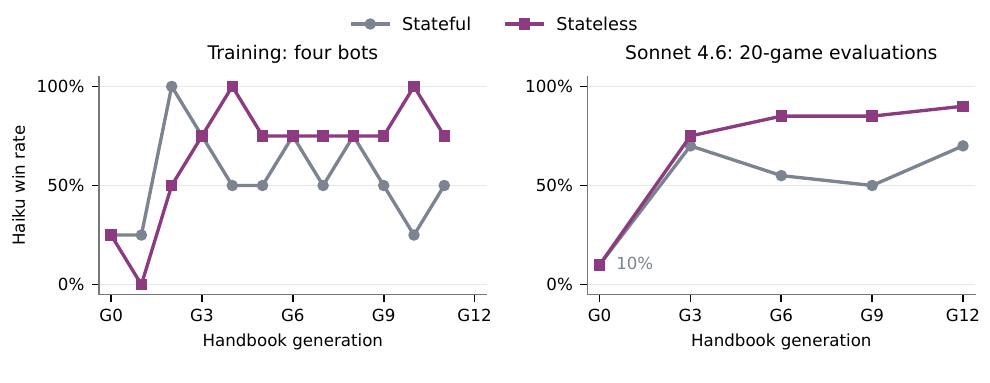}
\caption{\textbf{Rebuilding the handbook from scratch gives a steadier trajectory.} The stateful teacher improves early and then regresses; the stateless teacher's win rate against Sonnet~4.6 rises from 75\% at G3 to 90\% at G12. G0 is the shared baseline (2/20).}
\label{fig:skills-teacher}
\end{figure}

\begin{table}[!ht]
\centering\small
\caption{Records against Sonnet~4.6 over twenty games per checkpoint.}
\label{tab:skills-teacher}
\begin{tabular}{lrr}
\toprule Checkpoint & Stateful & Stateless\\\midrule
G0 & \multicolumn{2}{c}{2--18--0 (shared)} \\
G3 & 14--1--5 & 15--3--2 \\
G6 & 11--3--6 & 17--3--0 \\
G9 & 10--9--1 & 17--2--1 \\
G12 & 14--5--1 & 18--0--2 \\
\bottomrule
\end{tabular}
\end{table}

The stateful teacher anchors on its most recent games. After its first session, 265 of its 299 file reads (88.6\%) target the latest generation, and nine of its twelve rationales open by naming the preceding generation. Each of its handbooks shares 52\% of its vocabulary with the previous one, falling to 37\% six generations back, whereas stateless handbooks share 38--39\% at every distance. Over the 48 training games, the stateful and stateless runs record 26--10--12 and 32--11--5; the Economy bot stays hard for both, with two wins each.

\paragraph{Anonymized evidence.}
\label{app:anon}
The stateless teacher sees each game under a random name, with the scripted opponents relabeled A--D in random order. Results keep only the seat, outcome, final tick, map, and behavioral metrics; token counts and costs are removed. The student's own calls are included without the handbook it was given and with skill names redacted, and replays are excluded. Before every teacher session, a check confirms that no bot name, generation label, or cost field remains.

\subsection{Bot-trained transfer}
\label{app:skills-transfer}
Haiku and Luna each train for twelve generations against the four bots, one game per bot per generation, with the stateless teacher. Each evaluated handbook is frozen and plays twenty games per opponent, ten seeds in both seats; evaluation games never reach the teacher. Haiku plays without thinking and Luna with xhigh reasoning, their tournament configurations, and Sonnet, Sol, and Opus use high reasoning (Figure~\ref{fig:skills}). For panels (c) and (d) of that figure, each of the student's games is assigned to the nearest tournament profile of Appendix~\ref{app:swiss-tsne} and drawn in the tournament embedding; the large markers are the medians of a checkpoint's games.

\subsection{Haiku's thinking configuration}
\label{app:skills-haiku-thinking}
Haiku keeps its tournament configuration, without thinking or scratchpad, which won six of ten games against Haiku with a 1,024-token thinking budget (Appendix~\ref{app:thinking-results}). Table~\ref{tab:skills-haiku-thinking} evaluates the same frozen handbooks with Haiku's thinking off and on, against Sonnet~5 without thinking and with high thinking. At G12, enabling Haiku's thinking changes the result by one win in either direction: 19/20 to 20/20 against Sonnet without thinking, and 11/20 to 10/20 against Sonnet with high thinking.

\begin{table}[!htbp]
\centering\small
\setlength{\tabcolsep}{5pt}
\caption{\textbf{Haiku's G12 results are similar with and without thinking.} Haiku's wins--losses--draws over twenty games per cell, with the same handbook in all four conditions.}
\label{tab:skills-haiku-thinking}
\begin{tabular}{lrrrr}
\toprule
& \multicolumn{2}{c}{Sonnet: thinking off} & \multicolumn{2}{c}{Sonnet: high thinking}\\
\cmidrule(lr){2-3}\cmidrule(lr){4-5}
Checkpoint & Haiku off & Haiku on & Haiku off & Haiku on\\
\midrule
G3 & 14--6--0 (70\%) & 16--4--0 (80\%) & 5--15--0 (25\%) & 8--12--0 (40\%) \\
G6 & 18--2--0 (90\%) & 20--0--0 (100\%) & 4--16--0 (20\%) & 7--12--1 (35\%) \\
G9 & 20--0--0 (100\%) & 20--0--0 (100\%) & 7--12--1 (35\%) & 13--7--0 (65\%) \\
G12 & 19--1--0 (95\%) & 20--0--0 (100\%) & 11--9--0 (55\%) & 10--10--0 (50\%) \\
\bottomrule
\end{tabular}
\end{table}

\subsection{Training-opponent regimes}
\label{app:skills-regimes}
We compare five training conditions for Haiku and Luna: each of the four bots alone, and all four mixed, with one game against each per generation. Every generation gives the student four games on \texttt{simple\_2p}, two in each seat. Handbooks are written by the stateless teacher of Appendix~\ref{app:skills-teacher}, and there is one training run per student and condition.

We evaluate the frozen G3 and G6 handbooks against eight opponents without handbooks: Opus, Sol, Gemini~3.7 Flash, Sonnet, Gemini~3.1 Pro, Terra, Luna, and Haiku. Each opponent is played on the five maps in both seats, which gives ten games per opponent and 80 per handbook.

\begin{figure}[!ht]
\centering
\includegraphics[width=\linewidth]{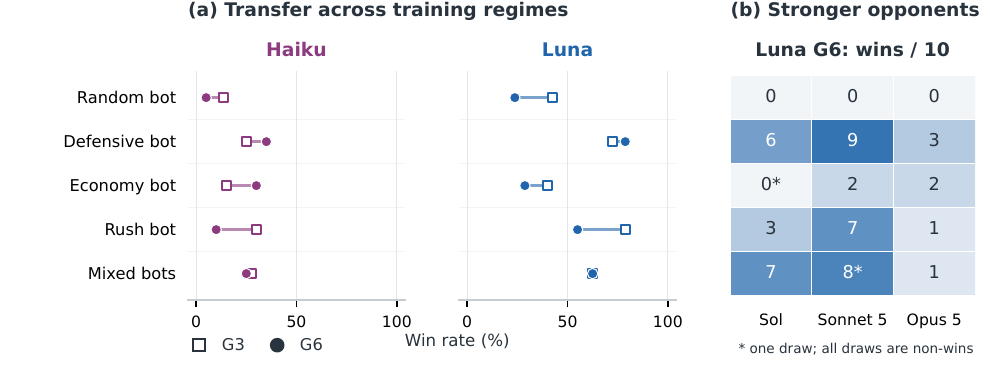}
\caption{\textbf{The training opponent decides which counter is distilled, and only some counters transfer.} (a) G3 and G6 win rates against the eight-opponent panel, 80 games each. (b) Luna's G6 wins out of ten against three stronger opponents.}
\label{fig:skills-regimes}
\end{figure}

The training bot changes the handbook and its results (Table~\ref{tab:skills-regimes-haiku}). At G6, Defensive-bot Luna wins 63/80 panel games, against 50/80 for mixed bots, 44/80 for Rush, 23/80 for Economy, and 19/80 for Random. Haiku's G6 records range from 4/80 with Random to 28/80 with Defensive.

\begin{table}[!htbp]
\centering\small
\setlength{\tabcolsep}{3.6pt}
\caption{\textbf{Results by training opponent.} Wins out of 80 against the panel and out of ten against each opponent. Each asterisk marks one drawn game. Flash and Pro are Gemini~3.7 Flash and Gemini~3.1 Pro.}
\label{tab:skills-regimes-haiku}
\label{tab:skills-regimes-luna}
\begin{tabular}{@{}llrrrrrrrrrr@{}}
\toprule Student & Training opponent & G & Panel & Opus & Sol & Flash & Sonnet & Pro & Terra & Luna & Haiku\\\midrule
Haiku & Random bot & G3 & 11\dr{1} & 0 & 0 & 0 & 2 & 1 & 0 & 1\dr{1} & 7 \\
 & & G6 & 4 & 0 & 0 & 0 & 0 & 0 & 0 & 0 & 4 \\
 & Defensive bot & G3 & 20 & 0 & 1 & 0 & 0 & 2 & 2 & 6 & 9 \\
 & & G6 & 28\dr{1} & 0 & 0 & 1\dr{1} & 7 & 2 & 4 & 5 & 9 \\
 & Economy bot & G3 & 12 & 0 & 1 & 0 & 0 & 0 & 1 & 2 & 8 \\
 & & G6 & 24 & 1 & 0 & 1 & 2 & 2 & 3 & 6 & 9 \\
 & Rush bot & G3 & 24 & 0 & 2 & 0 & 3 & 5 & 3 & 2 & 9 \\
 & & G6 & 8\dr{1} & 0 & 0 & 0 & 1 & 0 & 1 & 0\dr{1} & 6 \\
 & Mixed bots & G3 & 22 & 0 & 0 & 0 & 1 & 3 & 4 & 5 & 9 \\
 & & G6 & 20 & 0 & 2 & 0 & 2 & 2 & 3 & 3 & 8 \\
\midrule
Luna & Random bot & G3 & 34 & 0 & 5 & 0 & 3 & 5 & 2 & 9 & 10 \\
 & & G6 & 19\dr{1} & 0 & 0 & 0 & 0 & 4\dr{1} & 4 & 2 & 9 \\
 & Defensive bot & G3 & 58 & 3 & 6 & 3 & 8 & 8 & 10 & 10 & 10 \\
 & & G6 & 63 & 3 & 6 & 6 & 9 & 9 & 10 & 10 & 10 \\
 & Economy bot & G3 & 32\dr{1} & 0 & 3 & 1 & 0\dr{1} & 4 & 7 & 7 & 10 \\
 & & G6 & 23\dr{1} & 2 & 0\dr{1} & 1 & 2 & 1 & 2 & 6 & 9 \\
 & Rush bot & G3 & 63 & 2 & 8 & 4 & 10 & 9 & 10 & 10 & 10 \\
 & & G6 & 44 & 1 & 3 & 1 & 7 & 7 & 6 & 9 & 10 \\
 & Mixed bots & G3 & 50 & 2 & 6 & 3 & 7 & 7 & 8 & 7 & 10 \\
 & & G6 & 50\dr{3} & 1 & 7 & 5 & 8\dr{1} & 8 & 5\dr{2} & 7 & 9 \\
\bottomrule
\end{tabular}
\end{table}

\paragraph{What each regime teaches.}
Each handbook encodes a counter to what its training opponent fielded. For Luna:
\begin{itemize}[leftmargin=*,itemsep=2pt,topsep=3pt]
\item \textbf{Defensive bot.} This bot fields only soldiers. The handbook prescribes five workers, one barracks, one soldier, and then continuous marauders, and wins 63/80.
\item \textbf{Economy bot.} This bot fields only marauders. The handbook defaults to pure strikers, two barracks, and a 12--14-striker attack. Luna loses its first four training games against this bot and wins all twenty after that, but the counter does not transfer: in evaluation, 83\% of its basic-unit hits land in unfavorable matchups, and it wins 23/80.
\item \textbf{Rush bot.} From G2 onward Luna wins every training game, so the teacher sees only wins. Between G3 and G6 the handbook's worker cap drifts from 8--9 to 20--24. Mean living workers rise from 8.5 to 14.7, and the marauder share falls from 65.7\% to 37.0\%.
\item \textbf{Mixed bots.} The handbook has no default composition. It prescribes 7--8 workers before the first barracks, a second barracks around ticks 40--50, and production chosen after scouting, and wins 50/80 at both checkpoints.
\end{itemize}

\begin{table}[!htbp]
\centering\small
\caption{\textbf{Luna's G6 production and results by training opponent.} Shares are mean per-game percentages of the non-worker units produced. A10 is the share of the 80 evaluation games nearest that profile.}
\label{tab:skills-regimes-mechanisms}
\begin{tabular}{lrrrrr}
\toprule Training opponent & Soldier \% & Marauder \% & Striker \% & A10 \% & Wins / 80\\\midrule
Rush bot & 33.6 & 37.0 & 15.0 & 26.2 & 44\\
Defensive bot & 10.3 & 81.0 & 0.3 & 68.8 & 63\\
Economy bot & 0.0 & 0.0 & 99.0 & 1.2 & 23\\
\bottomrule
\end{tabular}
\end{table}

\begin{figure}[p]
\centering
\includegraphics[width=.80\linewidth,trim=0 80bp 0 0,clip]{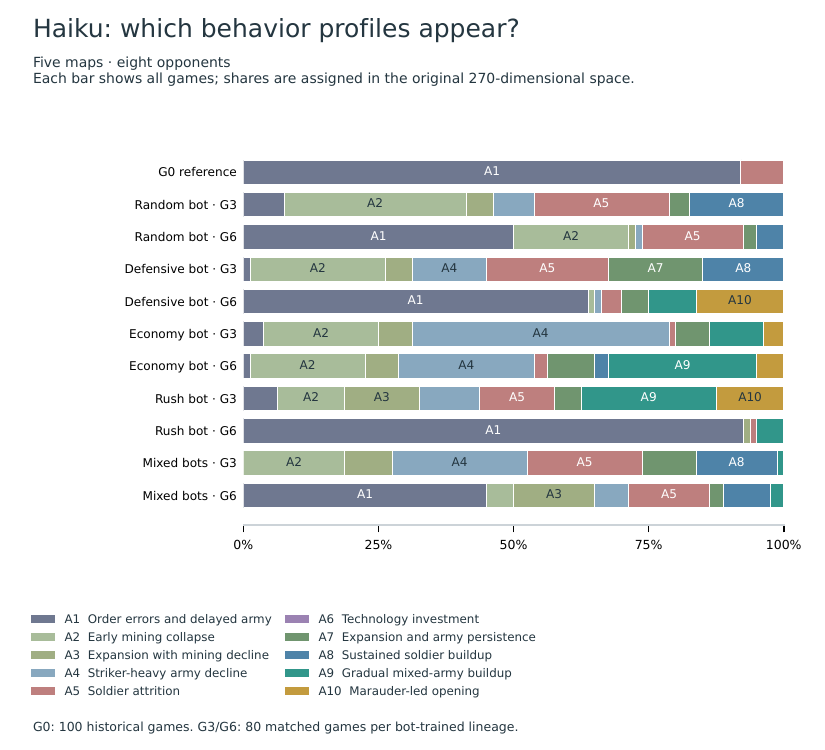}\\[4pt]
\includegraphics[width=.80\linewidth]{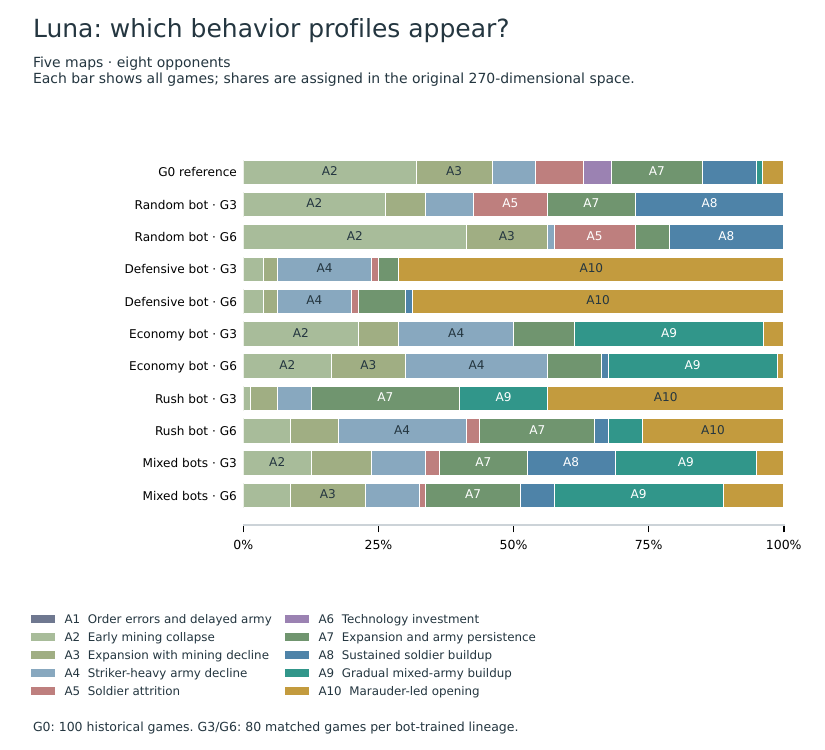}
\caption{\textbf{Profiles reached under each training opponent.} Share of evaluation games nearest each tournament profile, 80 games per training opponent and checkpoint. The G0 bar is the student's 100 tournament games.}
\label{fig:skills-haiku-profiles}
\label{fig:skills-luna-profiles}
\end{figure}

\paragraph{Behavioral profiles.}
\label{app:skills-behavior}
We assign each evaluation game to the nearest of the ten tournament profiles (Figure~\ref{fig:skills-haiku-profiles}). The same student reaches different profiles under different training opponents:
\begin{itemize}[leftmargin=*,itemsep=2pt,topsep=3pt]
\item \textbf{Defensive-bot Luna} lands in A10 in 71\% and 69\% of its G3 and G6 games, and wins 72.5\% and 78.8\% of its games.
\item \textbf{Mixed-bot Luna} most often lands in A9 (26\% and 31\%) and wins 62.5\% at both checkpoints.
\item \textbf{Mixed-bot Haiku} leaves A1, the profile of 92\% of its tournament games: 0\% at G3 and 45\% at G6.
\item \textbf{Economy-bot Luna}'s all-striker games resemble no tournament profile: 78--80\% lie beyond the distance that contains 95\% of tournament player-games.
\end{itemize}

\subsection{Experience versus rules-only guidance, and repeated training}
\label{app:skills-controls}
Does the gain come from game experience or from the teacher's own knowledge of the game, and does the procedure give the same result when repeated? We test both with Haiku against Sonnet~5 with high thinking, ten games per handbook (five seeds in both seats).
\begin{itemize}[leftmargin=*,itemsep=2pt,topsep=3pt]
\item \textbf{Rules only.} The teacher of the main experiment (GPT-5.6 Sol) writes a handbook under the same limits from the rules alone, with no access to any game.
\item \textbf{Re-distillation.} Three fresh teacher sessions each rewrite the G12 handbook from the same 48 training games as the original.
\item \textbf{Independent runs.} The whole twelve-generation procedure is repeated three times from scratch. Each run plays its own 48 training games with new seeds and has its own teacher sessions, here with GPT-6 Astra as teacher.
\end{itemize}

\begin{table}[!htbp]
\centering\small
\setlength{\tabcolsep}{8pt}
\caption{\textbf{Handbooks distilled from experience win; a rules-only handbook does not.} Haiku's wins out of ten against Sonnet~5 with high thinking, on the same ten seed and seat pairs for every handbook.}
\label{tab:skills-controls}
\begin{tabular}{@{}llr@{}}
\toprule
Handbook & Teacher's evidence & Wins / 10 \\
\midrule
Rules only & Rules, no games & 0 \\
\midrule
Original G12 & 48 training games & 3 \\
Re-distillation 1 & Same 48 games & 5 \\
Re-distillation 2 & Same 48 games & 4 \\
Re-distillation 3 & Same 48 games & 7 \\
\midrule
Independent run 1 & Its own 48 games & 6 \\
Independent run 2 & Its own 48 games & 5 \\
Independent run 3 & Its own 48 games & 5 \\
\bottomrule
\end{tabular}
\end{table}

The rules-only handbook wins no game, while the seven experience-derived handbooks win 35 of 70 (Table~\ref{tab:skills-controls}; two-sided Fisher exact test, $p=0.004$). The three independent runs win 16 of 30, in line with 19 of 40 for the handbooks written from the original games, and the seven handbooks do not differ detectably from one another (Freeman--Halton exact test, $p=0.72$). Against GPT-5.6 Luna with high reasoning, the original handbook and its three re-distillations win 10, 9, 10 and 9 of 10 games.

The independent runs also rediscover the same strategy. All seven experience-derived handbooks prescribe six to eight workers, a barracks started around ticks 17--25, and a default army of marauders, and they differ only in secondary advice. Three of the four training bots are deterministic, so every run faces the same opponents, and the teacher distills a plan that works against all of them. The rules-only handbook (printed below) instead covers every mechanic and leaves the choices to the student: a ``mixed core'' army, a second barracks ``only if income can feed both'', and expansion ``only when rich and secure''.

\clearpage
\subsection{Example handbooks}
\label{app:skills-handbook}
Below are the complete G12 handbooks of Haiku and Luna from the mixed-bot runs, and the rules-only handbook of Appendix~\ref{app:skills-controls}. Haiku's fixes a marauder recipe and spends a skill on order validity; the rules-only handbook hedges on every choice; Luna's conditions production on the visible enemy army.
\begin{skillhandbook}{Skill Handbook Prompt: Haiku G12}
These are lessons distilled from previous TextCraft games. They are advice, not rules. Apply them when relevant, but obey the game rules and current state above all.\par\medskip
\textbf{skill\_opening\_economy:} Immediately build an extractor on the visible starting resource beside your command center and queue four workers from the command center. Wait for completion, then harvest with every unloaded idle worker, including the builder and new workers. Harvest handles all travel/pickup/return; never move workers onto the extractor or command-center tile. Reach 7--8 total workers, then stop. Keep 6--7 harvesting while one builds. If that gives much less than about 24--32 income, find unloaded workers on idle/hold/stale move and reissue harvest. Do not redirect loaded workers returning to deposit.\par\smallskip
\textbf{skill\_timely\_barracks\_and\_second\_production:} After the opening queue, save 150 and start a barracks around ticks 17--21; starts around 29--51 repeatedly failed. Use a currently visible empty passable tile near home, not a resource, structure, ramp, or shuttle lane. Make exactly one scout, then continuously queue marauders. Prefer army to workers beyond the 7--8 cap. If safe and 150 remains after an initial unit queue, add a second barracks in early-midgame; this won the hard marauder mirror. Resume a stalled barracks at the identical site/type. Delay tech, upgrades, and expansion until a strong tier-1 army exists.\par\smallskip
\textbf{skill\_composition\_rules:} Default to a marauder core. Correct matchup facts: marauders get +40\% versus soldiers, while soldiers deal −40\% to marauders. Soldiers are only for enemy striker masses (+30\% versus strikers); never make soldiers into marauders. Strikers are modifier-favored against marauders but must close range, so at most use a small screen, not a replacement for ranged marauders. In a marauder mirror, rely on two-barracks production, cohesion, and numbers rather than panic-pivoting. One scout supplies vision and should not duel full-health combat units.\par\smallskip
\textbf{skill\_scout\_and\_mass:} Send the scout through passable terrain toward the unexplored frontier away from home. After finding the enemy, pull it behind the marauders to preserve vision. Gather reinforcements at one safe forward staging area instead of trickling them. Use 4--6 marauders against an undefended, worker-heavy base; against a marauder mass, consolidate about 8--10 or a clear numerical edge. If intel shows many workers or an extra command center but little army, attack with the first coherent force rather than matching the greed.\par\smallskip
\textbf{skill\_focused\_attacks:} Units fire only with attack orders. When an enemy is visible, use exact LIVE INTEL IDs, not move/hold. Assign about 3--4 marauders per enemy instead of one-on-one splitting; use everyone on the sole target. Kill combat units before workers and remove soldiers quickly. Attack pursuit already closes and may fire that tick, so do not alternate attack with move. Valid attacks persist through cooldown. On ENEMY\_KILLED, immediately give those units a new visible target because they will not auto-select. Never invent an ID or give one unit multiple orders in a tick.\par\smallskip
\textbf{skill\_high\_ground\_defense:} If a large force approaches or a base structure is attacked, stop the outbound march and reunite the army between the threat and command center. Prefer the visible high-ground side of a ramp: firing down gives +25\%, charging uphill is penalized. Keep the scout behind the line, then focus the closest visible attackers. Do not chase one retreating unit onto low ground while the main army advances. Keep workers harvesting and producing reinforcements; do not use them as combat scouts.\par\smallskip
\textbf{skill\_finish\_decisively:} After the enemy army breaks, advance the whole core with scout vision. Destroy a producing barracks first if it can replenish defenders; otherwise attack an exposed command center directly. Then keep every combat unit on the exact visible command-center ID until it dies. Ignore extractors and scattered workers. Retarget after each destruction alert. If the base is in fog, move toward its last observed area, reveal it, then use exact IDs. Convert the advantage before resources deplete instead of waiting for a larger army or timing out.\par\smallskip
\textbf{skill\_order\_validity\_check:} Before acting, read ALERTS, resources, queues, and exact IDs. Command center produces workers; barracks produces tier-1 combat units. Ensure all new orders together fit the bank, and do not blindly add to an existing queue. Build only on a currently visible empty legal tile; harvest only from your completed owned extractor with real worker IDs. If ORDER\_FAILED appears, fix the cause instead of repeating it. Orders persist, so do not reissue them every tick; when nothing needs changing, return [].\par\smallskip
\end{skillhandbook}
\begin{skillhandbook}{Skill Handbook Prompt: Rules only (no game experience)}
These are lessons distilled from previous TextCraft games. They are advice, not rules. Apply them when relevant, but obey the game rules and current state above all.\par\medskip
\textbf{skill\_turn\_loop:} Each tick read alerts, resources/queues, idle or damaged workers, cooldowns, and LIVE INTEL. Preserve useful persistent harvest/attack orders. Give newly idle fighters a purpose. Visible enemy means issue attacks using exact current IDs; units never auto-attack. Retarget when a target dies. If no ID is visible, move to gain vision---never invent one. Issue only affordable, legal JSON orders.\par\smallskip
\textbf{skill\_opening\_economy:} Prioritize income: build an extractor on a visible nearby resource tile that a worker can approach on matching elevation, and produce a few workers if affordable. When complete, assign roughly 3--5 workers to harvest, adjusting for travel and congestion. Save next for a barracks. Usually use one builder; extra builders speed urgent construction but stop mining. Replace workers and redirect them when nodes deplete. Add another safe extractor when income supports production, preferring short routes to a command center. Expand with another CC only when rich and secure.\par\smallskip
\textbf{skill\_build\_order:} Use extractor/workers $\rightarrow$ barracks $\rightarrow$ scout $\rightarrow$ continuous tier-1 army. Add a second barracks only if income can feed both. Do not buy tech while too weak to survive pressure. With a stable army, machine\_shop is a useful first branch: ordnance for combat, boosters for pursuit and positioning. Add tech\_lab later; plating affects only future units. Tanks and enforcers supplement rather than replace tier-1 units. Delay weapons\_lab/orbitals until economy and map control are strong.\par\smallskip
\textbf{skill\_scouting\_positioning:} Send an early scout through unexplored frontiers, ramps, and high-ground approaches while avoiding fights. Discover enemy economy, tech, army mix, and valid IDs. Keep a scout near the army for targeting vision. Stage forces on reachable high ground or atop a ramp between threats and your economy, with a retreat route. Move toward remembered positions to reacquire enemies, but attack only entities currently in LIVE INTEL.\par\smallskip
\textbf{skill\_combat\_micro:} Fight together, not in a trickle. Focus several units on one killable threat, then retarget immediately; prioritize fragile high-impact tanks/marauders and units countering your army before workers or structures. Keep marauders/tanks behind soldiers and send strikers around onto marauders/tanks. Use attack pursuit when the target is visible because it can move and fire in one tick. Favor high ground and avoid uphill fights unless overwhelming. Reposition during cooldown only for real safety/range gain. Spread against splash; rush inside a tank's minimum range with fast units while ranged units hit from outside. Do not chase one unit deep into fog.\par\smallskip
\textbf{skill\_composition\_rules:} Default to a mixed core: soldiers in front, marauders at range, one scout for vision, and some strikers for flanks. Adapt to observed forces: soldiers counter strikers; marauders counter soldiers; strikers counter marauders. Against tanks, spread and close with strikers; against enforcers, kite and focus with ranged units. Scouts are information units, not the main army. When enemy composition is unknown, prefer soldiers plus marauders rather than one unit type.\par\smallskip
\textbf{skill\_defense\_recovery:} When raided, stop distant chases, gather the army between attackers and workers, gain vision, and focus exact IDs. Move workers away only when threatened; otherwise preserve income. Save the CC and production before an exposed/depleted extractor. After combat, a few workers may repair completed key structures, but stop before repair costs starve production. Resume an interrupted build with the same type and site at no new cost. If behind, make cheap counters and defend a ramp/high ground instead of teching. Counterattack only after the threat is removed.\par\smallskip
\textbf{skill\_orbital\_endgame:} An ORBITAL\_STRIKE alert is urgent: immediately move every unit/worker outside the targeted 3$\times$3; do not wait for the countdown or cluster at its edge. If launching one, spend 300 only when secure and target a dense force or valuable immobile structures unlikely to escape during the warning. Late game, keep scouting and replacing depleted income. After winning a fight, destroy workers/extractors and production, then concentrate on the command center while continuing unit production and retaining raid defense.\par\smallskip
\end{skillhandbook}
\begin{skillhandbook}{Skill Handbook Prompt: Luna G12}
These are lessons distilled from previous TextCraft games. They are advice, not rules. Apply them when relevant, but obey the game rules and current state above all.\par\medskip
\textbf{skill\_opening\_script:} Default opening: on tick 0, have one worker build an extractor on the visible resource nearest the command center and have the command center produce a worker in the same response. Keep the other worker home; never scout across the map with an opening worker. When the extractor completes, order all workers to harvest it. Keep producing workers and assign each new one to harvest. At about 8 workers, build one barracks on a visible empty passable tile near home without blocking the ramp or extractor. Use one builder; continue to 12--14 workers.\par\smallskip
\textbf{skill\_worker\_cap\_and\_two\_barracks:} Treat 12--14 workers as a hard standard cap; then stop workers and spend on fighters. From the first barracks make one scout and 1--2 fighters, then build a second barracks when 150 is available and the first is producing---usually in the late 30s to 50, not after losing the army. If resources bank above 150 with only one barracks, build the second. Keep both barracks busy. Skip tech, upgrades, and a second command center during the standard fight; successful games were decided by barracks armies before these paid back.\par\smallskip
\textbf{skill\_scout\_without\_sacrificing\_economy:} Use one scout, not a worker, for early intel. Move in short legs along passable terrain toward unexplored routes and enemy approaches; inspect the minimap so resources, walls, and edges do not cause no-path orders. Reveal the enemy army, base, and exact LIVE INTEL IDs, but do not fight. Withdraw or circle for vision when threatened. Do not mass scouts. Rally later fighters safely near the barracks and move them forward in groups.\par\smallskip
\textbf{skill\_counter\_the\_visible\_army:} Counter the most numerous dangerous type in LIVE INTEL. Against marauders, mass strikers and avoid soldiers. Against soldiers, mass marauders and avoid strikers. Against strikers, mass soldiers. For a mixed army, split production by visible counts while keeping a majority counter to its majority. While composition is unknown, one scout plus 1--2 soldiers is enough; switch later cycles when intel arrives. Do not cancel a fighter already in production just to switch; finish it, then change the next cycle and avoid the 25\% cancellation loss.\par\smallskip
\textbf{skill\_assemble\_waves\_and\_use\_terrain:} Do not send each new fighter alone. Gather reinforcements at a safe staging tile on the route, then move a wave of roughly 5--8 fighters together; attack earlier only to defend or when the visible enemy is clearly weaker. If a small group meets a larger force, pull it toward reinforcements. When available, stage ranged units on \texttt{\^{}} high ground or defend the top of a ramp. Keep marauders behind tougher units and let strikers close together.\par\smallskip
\textbf{skill\_explicit\_focus\_fire:} Units never auto-attack. When a target is in LIVE INTEL, issue \texttt{attack} with its exact ID to every suitable fighter. Attack pursuit can move and fire that tick; normal movement prevents firing. Focus one enemy fighter at a time, preferring low HP or a counter to your army, then retarget immediately when it dies. Do not split attacks onto workers or structures while enemy weapons still fire. Give each unit at most one order per tick; do not move a ready ranged unit that can attack. An accepted attack can continue through fog, but a dead target leaves units needing a new order.\par\smallskip
\textbf{skill\_defend\_then\_finish\_cleanly:} A base-under-attack alert overrides a distant raid: recall the army, produce counters from every idle barracks, and attack visible invaders together. Workers cannot fight; repair the command center with only 1--2 workers after the army engages and only while resources remain. Ignore enemy workers until armed units are handled. At the enemy base, kill active fighters first. Once its army breaks, focus the whole army on the visible command center instead of chasing workers or splitting across structures.\par\smallskip
\textbf{skill\_every\_tick\_checklist:} Each tick check: (1) alerts and LIVE INTEL; (2) command center if below the worker cap; (3) every idle barracks; (4) new workers needing \texttt{harvest}; (5) fighters with dead targets or no useful order. Use only current owned IDs or exact LIVE INTEL target IDs. Build only on currently visible valid cells and issue at most one order per unit. After RESOURCE\_DEPLETED, stop harvest retries and worker production; spend the stockpile on fighters and make the grouped finishing push. Idle barracks and large banks are emergencies, not reasons to save for tech.\par\smallskip
\end{skillhandbook}

\FloatBarrier
\section{2v2 Team Games}
\label{app:team-tsne}

\subsection{Setup}
\label{app:team-protocol}
Each of three rosters plays 24 matches in which a chatting copy of the roster faces a silent copy: beginner (Haiku 4.5 with DeepSeek V4 Flash), intermediate (Sonnet 5 with Gemini 3.1 Pro), and expert (Astra with Fable 5.1). The 24 matches cover the three four-player maps, both sides of the map, and the four orders of the players within the two teams. Observations are private to each player. The 72 games contain 9,919 ticks, 37,531 model turns, and 17,276 messages.

\subsection{Team profiles}
\label{app:team-profiles}
We describe each team in each game, 144 team-games, with the features of Appendix~\ref{app:swiss-clustering} computed over both teammates, add 17 features that measure coordination, and group the team-games into ten team profiles, B1--B10 (Figure~\ref{fig:team-tsne-team}, Table~\ref{tab:team-tsne-team-temporal}; procedure in Appendix~\ref{app:team-method}). The statistics that mark each profile are:

\begin{description}[leftmargin=1.2em,itemsep=2pt,topsep=4pt]
\item[B1, expansion with mining collapse.] All 18 games have an unprotected base under pressure, and in 17 the army is one-sided for a sustained period.
\item[B2, high mixed-army production and fighting.] Military production averages 20.7 resources per tick, and the teammates repeatedly fight together in 13 of 14 games.
\item[B3, persistent two-player pressure.] All 16 teams expand, and both teammates have an army for 74.2\% of the game.
\item[B4, soldier pressure after income falls.] Soldiers are 91.2\% of the non-worker units produced, and all 18 games develop sustained forward pressure.
\item[B5, declining armies and exposed bases.] 12 of 14 teams expand, but all 14 have an unprotected base under pressure.
\item[B6, late, one-sided armies and order errors.] The first military unit appears at tick 52.7 on average, and both teammates have an army for only 16.8\% of the game.
\item[B7, scout opening to sustained pressure.] Both teammates have an army for 49.7\% of the game, and all 14 games develop sustained forward pressure.
\item[B8, busy production with fading support.] Barracks are busy 79.5\% of the time, yet in all 13 games the army is one-sided for a sustained period.
\item[B9, sustained soldier reinforcement.] Barracks are busy 69.1\% of the time, and the teammates repeatedly fight together in all 18 games.
\item[B10, single-game tank-supported assault.] One game, in which tanks are 16.7\% of the non-worker units produced.
\end{description}

\begin{figure}[!htbp]
\centering
\includegraphics[width=\linewidth]{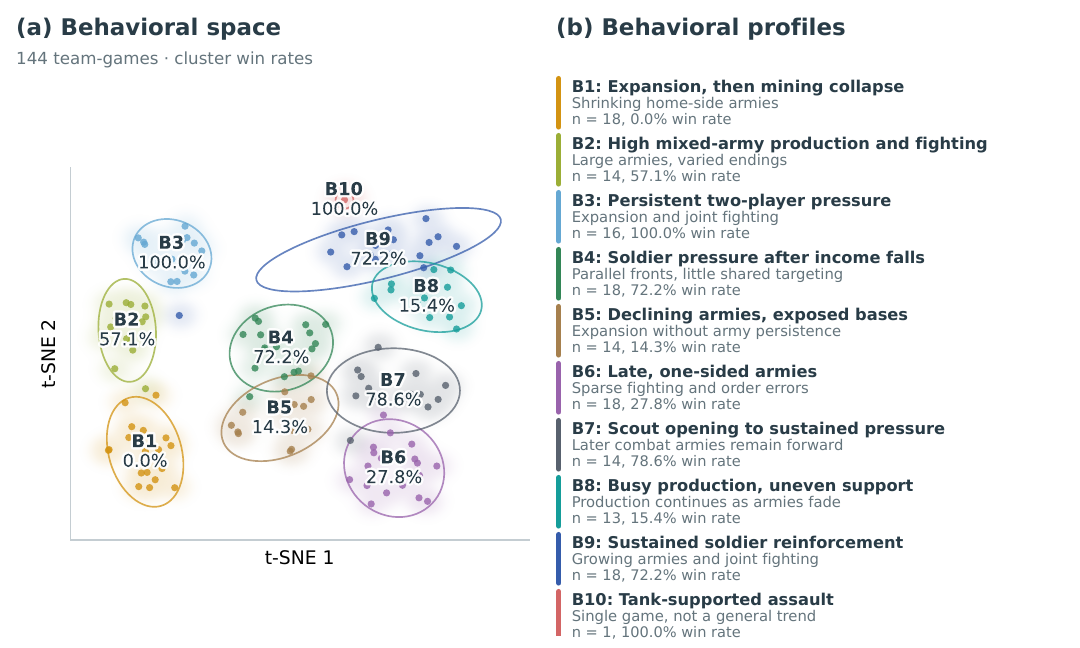}
\caption{\textbf{Team profiles.} (a) The 144 team-games, colored by profile, with the win rate of each profile. (b) Profile names and sizes.}
\label{fig:team-tsne-team}
\end{figure}

\begin{table}[!htbp]
\centering\footnotesize
\setlength{\tabcolsep}{3pt}
\renewcommand{\arraystretch}{1.12}
\caption{Team profiles, as in Table~\ref{tab:swiss-temporal-profiles} with the two teammates' economies and armies summed.}
\label{tab:team-tsne-team-temporal}
\begin{tabular}{@{}c>{\centering\arraybackslash}p{0.21\linewidth}cr@{\,$\to$\,}lr@{\,$\to$\,}lr@{\,$\to$\,}lc@{}}
\toprule Profile & Name & \shortstack{Team-\\games} & \multicolumn{2}{c}{Mining} & \multicolumn{2}{c}{Army value} & \multicolumn{2}{c}{Forward (\%)} & Win (\%) \\
 & & & 21--40 & 101--120 & 41--60 & 101--120 & 41--60 & 101--120 & \\
\midrule
B1 & Expansion with mining collapse & 18 & 34.9 & 0.8 & 260.6 & 86.1 & 12.5 & 8.3 & 0.0 \\
B2 & High mixed-army production and fighting & 14 & 36.8 & 17.9 & 465.6 & 669.8 & 27.4 & 62.1 & 57.1 \\
B3 & Persistent two-player pressure & 16 & 35.3 & 9.0 & 394.4 & 477.9 & 33.3 & 75.5 & 100.0 \\
B4 & Soldier pressure after income falls & 18 & 32.2 & 1.4 & 475.3 & 324.3 & 23.0 & 83.1 & 72.2 \\
B5 & Declining armies and exposed bases & 14 & 34.2 & 2.5 & 357.1 & 124.0 & 20.3 & 20.8 & 14.3 \\
B6 & Late, one-sided armies and order errors & 18 & 28.8 & 2.3 & 168.7 & 140.1 & 21.7 & 16.7 & 27.8 \\
B7 & Scout opening to sustained pressure & 14 & 32.1 & 4.4 & 258.0 & 366.2 & 21.2 & 87.5 & 78.6 \\
B8 & Busy production with fading support & 13 & 29.3 & 19.7 & 262.9 & 306.1 & 35.1 & 25.5 & 15.4 \\
B9 & Sustained soldier reinforcement & 18 & 32.1 & 25.7 & 354.0 & 892.4 & 23.7 & 53.9 & 72.2 \\
B10 & Single-game tank-supported assault & 1 & 27.2 & 35.6 & 362.0 & 1083.1 & 42.2 & 45.0 & 100.0 \\
\bottomrule
\end{tabular}
\end{table}

\begin{figure}[!htbp]
\centering
\includegraphics[width=\linewidth]{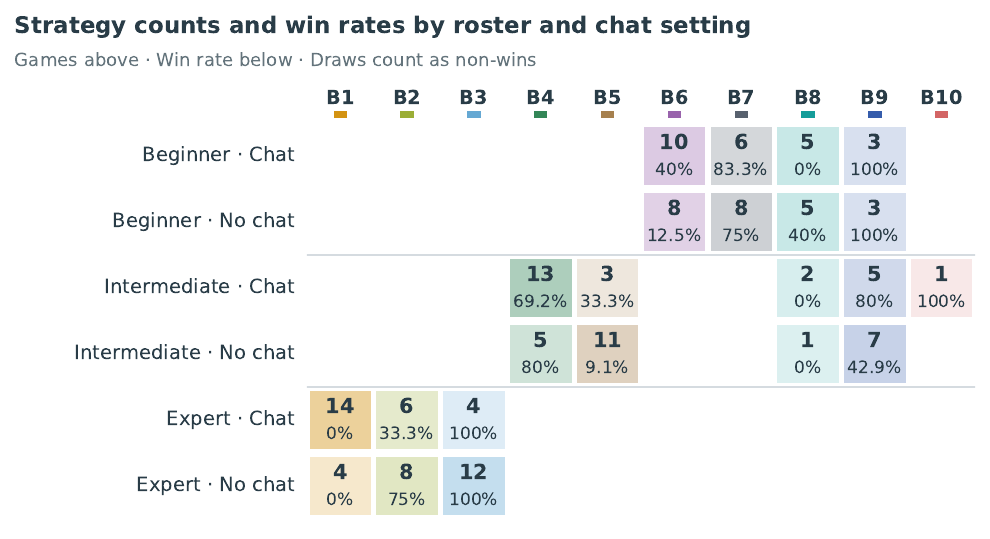}
\caption{\textbf{Team profiles by roster and chat setting.} Each cell gives the number of team-games in a profile (top) and their win rate (bottom); each row is 24 games.}
\label{fig:team-tsne-roster-shares}
\end{figure}

\subsection{Communication, coordination, and outcomes}
\label{app:team-chat-behavior}

Table~\ref{tab:team-tsne-chat-paired} subtracts the silent team's value from the chatting team's in each match and averages over a roster's matches. \emph{Forward} is the share of the army beyond the midpoint, \emph{joint army} the share of ticks in which both teammates have an army, \emph{joint combat} the share in which both fire, \emph{same target} the share of those in which both fire at the same tile, and \emph{imbalance} the relative difference between the teammates' army values.

\begin{table}[!htbp]
\centering\small
\setlength{\tabcolsep}{4pt}
\caption{Chat minus no chat: mean difference in percentage points, with the number of matches in parentheses.}
\label{tab:team-tsne-chat-paired}
\begin{tabular}{@{}lrrrrr@{}}
\toprule Roster & Forward & Joint army & Joint combat & Same target & Imbalance \\
\midrule
Beginner & -6.7 (24) & +4.3 (24) & -4.0 (24) & +7.5 (13) & -6.1 (24) \\
Intermediate & +15.7 (24) & +11.9 (24) & +8.4 (24) & +13.5 (24) & -14.9 (24) \\
Expert & -28.5 (24) & -11.4 (24) & -10.2 (24) & -8.0 (22) & +11.7 (24) \\
\bottomrule
\end{tabular}
\end{table}

The conversations show why the same channel has different effects across rosters. The examples link messages to the orders and replay states that followed; B labels give the chatting team's profile.

\paragraph{Beginner: waiting for an army that is not ready.}
For Haiku and DeepSeek Flash, the recurring problem is unreliable information about readiness. In the \href{https://textcraft-rts.com/\#game/50f1075964a01bb38990}{Simple Map Control loss} (beginner game 17, B6), Haiku says ``Hold position, don't overextend'' at tick 37. DeepSeek Flash replies ``Will hold till you join'' and redirects two soldiers to a staging position. But Haiku's repeatedly promised reinforcements do not exist yet: its first military unit appears at tick 66. In the \href{https://textcraft-rts.com/\#game/75b3fbd562464a912f3c}{Simple loss} (beginner game 15, B8), the mismatch persists during an attack. DeepSeek Flash reports ``7 soldiers engaging. Need your soldiers ASAP!'' while Haiku claims its barracks is complete, six ticks before it is.

Chat works better when both players can act on the plan. In the \href{https://textcraft-rts.com/\#game/cfc01a3761c5ad91788e}{King of the Hill win} (beginner game 20, B7), both field military by tick 40. At tick 51, DeepSeek Flash says ``Moving 3 soldiers to (20,15) to mass with you. Producing another.'' The reinforcement arrives and the pressure is sustained. Waiting and defensive regrouping explain why the chatting beginner team has better army balance but less forward presence and joint combat than its silent counterpart, and no advantage in wins.

\paragraph{Intermediate: requests become useful reinforcements.}
Sonnet and Gemini Pro more often talk about forces that already exist. In the \href{https://textcraft-rts.com/\#game/28257e4adb94c61a2ba3}{Simple win} (intermediate game 12, B9), Gemini Pro asks ``Push your army to (15, 15)!'' at tick 41. Instead of waiting for its preferred five soldiers, Sonnet sends four at tick 43, then the fifth while producing another.

Useful coordination also means dividing the work. In the \href{https://textcraft-rts.com/\#game/0f2f2fd06a0b6538aecc}{Simple Map Control win} (intermediate game 14, B4), Sonnet reports depleted income but keeps its soldiers active. It then spots undefended workers and redirects to the other opponent. Gemini Pro responds ``good call on destroying P2's economy'' and maintains its own attack. The team wins without ever firing at the same target.

The contrast is the \href{https://textcraft-rts.com/\#game/0b97b594d3ef05b37fe5}{King of the Hill loss} (intermediate game 23, B5). Gemini Pro reports being ``heavily outnumbered.'' Sonnet sends help, but diverts the first reinforcements to enemy workers while its ally's soldiers are dying, and Gemini Pro's army and base fall. Across the roster, chat more often sustains both players' participation than produces this mismatch, which is the higher forward presence and joint combat of Table~\ref{tab:team-tsne-chat-paired}.

\paragraph{Expert: a shared plan can weaken both armies.}
Astra and Fable exchange precise tactical information, but agreement can lock in an expensive plan. In the \href{https://textcraft-rts.com/\#game/177c6c8feeafd00545c0}{King of the Hill loss} (expert game 24, B1), they endorse saving for additional command centers while their fighting forces fall behind. Neither proposed command center starts. Later, Fable recommends defending separate plateaus, and Astra agrees to ``stopping all pursuit''. By tick 50, expert chat teams have produced less military than their silent opponents in 22 of 24 games.

They can also coordinate an assault successfully and still lose the game. In the \href{https://textcraft-rts.com/\#game/4b6620f50801d5e1df47}{Simple loss} (expert game 12, B2), the players regroup marauders and share target assignments. Their combined attack destroys an enemy command center, but Fable warns ``my base is lost, so your CC is the anchor.'' Its own base then falls, followed by Astra's.

The \href{https://textcraft-rts.com/\#game/06bf8338c817489eae55}{Simple Map Control win} (expert game 18, B3) shows the useful adjustment. Fable keeps producing while Astra saves, and Astra eventually announces ``switching savings to marauder reinforcements''. They keep attacking while exchanging updates on home defense. Chat helps when it supports timely military commitments and hurts when it prolongs an investment or an attack that leaves an ally unable to hold.

\subsection{Intervening on the agreed plan, message accuracy, and same-tick chat}
\label{app:team-chat-followups}

\paragraph{Intervention on the agreement.}
\label{app:team-chat-intervention}
We test whether the agreement causes the under-production in the setting where the link is strongest: the expert roster on King of the Hill, where the chatting team planned an additional command center in all eight games and lost all eight. We replay three of these games with their original seeds and seats and one change. A language model reads the team conversation at every tick. When a player proposes saving for or building an additional command center, its teammate's next message is edited before delivery: an objection is put first and any endorsement of the expansion is removed, while the rest of the message is kept. For example, at tick~32 of the first game Fable writes ``I'll save for a CC on mid hill near (12,17) for the hill minerals.'' Astra's reply, ``Securing NW hill with two marauders; I also need my own hill CC because workers cannot deposit at allied CCs.'', is delivered as ``P2, do not save for or build a hill CC. Securing NW hill with two marauders.'' In total 16 messages are edited (nine, three, and four per game), all between ticks 30 and 50. The players' orders and the silent team are not touched.

\begin{table}[!htbp]
\centering\small
\setlength{\tabcolsep}{5pt}
\caption{\textbf{Removing the agreement on the expansion restores production and reverses the outcome.} Expert roster on King of the Hill, chatting team against silent team. Military is the mean unit cost produced by tick~50, chatting / silent team; unspent is the chatting team's mean bank at tick~50.}
\label{tab:teamchat-intervention}
\begin{tabular}{@{}lrrrr@{}}
\toprule
Condition & Games & Chat W--L & Military by tick 50 & Unspent \\
\midrule
Next-tick chat, unedited & 8 & 0--8 & 320 / 541 & 334 \\
Reply replaced by an objection & 3 & 3--0 & 547 / 510 & 170 \\
\bottomrule
\end{tabular}
\end{table}

The chatting team wins all three games, against none of eight without the intervention (Table~\ref{tab:teamchat-intervention}; two-sided Fisher exact test, $p=0.006$). Home minerals run out around tick~60 on this map, so most of a team's army is built before then. Without the intervention, the chatting team has produced far less military than the silent team by tick~50 and holds twice the unspent resources; with it, the chatting team produces as much as its opponent. Once the expansion plan is gone, the teammates use the channel to fight together: they call targets by unit ID, report enemies the other cannot see, and time joint volleys (``let's kill their marauder together, not feed soldiers to it''). Both teammates had targeted 17 of the 56 enemy combat units the chatting team killed (silent team: 1 of 24), and the chatting team destroyed twice as much military as it lost (2,840 vs.\ 1,380 in unit cost).

\paragraph{Messages checked against the game state.}
Every message is linked to the sender's observation and to the replay, so claims can be checked. We select unqualified claims that a barracks is ready (``Barracks ready.'', ``Barracks done;'') and compare each with the observation before the message and the replay state after it. For Haiku, 21 of 25 such claims, spread over 15 games, had no completed barracks in either state. No false claim appears among DeepSeek Flash's 10, Sonnet's 25, Gemini Pro's 16, Astra's 13, or Fable's 20. The expert team's problem is the plan and not the reports: an early plan to save for an extra command center appears in 15 of its 24 chat games, yet across all 24 games only two extra command centers are started and one is completed, while the team holds more unspent resources at tick~50 than its silent counterpart (230 vs.\ 147 on average).

\paragraph{Same-tick communication.}
In the main experiment each player makes one model call per tick, which returns both orders and a message, so a message is read on the next tick. In a variant, each teammate first sends one message, and both then choose their orders with the full exchange in context, at the cost of a second model call per tick. For the expert roster on King of the Hill, the chatting team wins 5 of 8 games in this setting, against 0 of 8 with next-tick delivery. In the replays, models revise their orders within the tick, for example redirecting soldiers to an ally's threatened extractor after a request for help.

\subsection{How the team profiles are computed}
\label{app:team-method}
Team profiles follow the procedure of Appendix~\ref{app:swiss-clustering} on the 144 team-games, with every count pooled over the two teammates. To the 57 full-game features we add 17:
\begin{itemize}[leftmargin=*,itemsep=2pt,topsep=3pt]
\item \textbf{Opening milestones (3).} Ticks of the first extractor, first barracks, and first military unit, averaged over the teammates.
\item \textbf{Coordination (11).} Share of ticks with both armies alive; gap between the teammates' first military units; imbalance of their military production and of their fielded armies; difference in unit composition; proximity of the two armies; ticks in which both fire, at the same tile, and at the same opponent; a threatened base defended by the teammate, and a threatened base with no defender.
\item \textbf{Terrain (3).} Downhill fire that covers the teammate's base; both teammates firing downhill; both reaching the same plateau in an assault.
\end{itemize}
The result is 287 columns, clustered with $k=10$. B1--B10 are numbered along the shortest path through the profile centers in Figure~\ref{fig:team-tsne-team}(a), so consecutive labels are neighbors.

\end{document}